\documentclass[sigconf]{acmart}

\renewcommand\footnotetextcopyrightpermission[1]{}
\setcopyright{none}

\usepackage[utf8]{inputenc} 
\usepackage[T1]{fontenc}    
\usepackage{hyperref}       
\usepackage{url}            
\usepackage{booktabs}       
\usepackage{amsfonts}       
\usepackage{nicefrac}       
\usepackage{microtype}      
\usepackage{xcolor}         
\usepackage{graphicx}
\usepackage{float}
\usepackage{amsmath}
\usepackage{multirow}
\usepackage{xcolor}
\usepackage[many]{tcolorbox}
\usepackage{algorithm}
\usepackage{algpseudocode}
\usepackage{multirow}
\usepackage{makecell}
\usepackage{graphicx}
\usepackage{subcaption}
\algnewcommand\Input{\item[\textbf{Input:}]}
\algnewcommand\Output{\item[\textbf{Output:}]}

\AtBeginDocument{%
  }

\begin{document}

\title{BDIP-Net: Dual-Interaction Graph Learning for Property Prediction of  Bilayer Materials}

\author{An Vuong}
\affiliation{%
  \institution{University of Arkansas}
  \city{Fayetteville}
  \country{USA}}
\email{anv@uark.edu}

\author{Chen Zhao}
\affiliation{%
  \institution{Baylor University}
  \city{Waco}
  \country{USA}}
\email{chen\_zhao@baylor.edu}

\author{Jin Hu}
\affiliation{%
  \institution{University of Arkansas}
  \city{Fayetteville}
  \country{USA}}
\email{jinhu@uark.edu}

\author{Shui-Qing Yu}
\affiliation{%
  \institution{University of Arkansas}
  \city{Fayetteville}
  \country{USA}}
\email{syu@uark.edu}

\author{Xintao Wu}
\affiliation{%
  \institution{University of Arkansas}
  \city{Fayetteville}
  \country{USA}}
\email{xintaowu@uark.edu}



\renewcommand{\shortauthors}{Vuong et al.}

\begin{abstract}
 Stacked bilayer materials exhibit rich stacking-dependent properties driven by the interplay between strong intra-layer bonding and weak inter-layer van der Waals interactions. The computational discovery of such materials is challenging because accurate structure generation typically relies on expensive DFT-based optimization, while existing machine-learning models often fail to explicitly distinguish different interaction types during property prediction. To address these challenges, we propose a machine-learning framework for efficient construction and property prediction of stacked bilayer materials. The framework employs a MatterSim-D3-based structural optimization workflow to generate DFT-quality bilayer structures from monolayer building blocks and stacking configurations at substantially reduced computational cost. For property prediction, we introduce BDIP-Net (Bilayer Dual-Interaction Potential Network), a graph neural network that explicitly models intra-layer and inter-layer interactions through interaction-specific potential representations and adaptive message fusion. We evaluate the proposed framework on BiDB, HetDB, and SAMBA, encompassing homobilayers, heterobilayers, and twisted bilayer systems. Results show that the MatterSim-D3-based workflow closely reproduces DFT-PBE-D3 optimized structures, while BDIP-Net consistently outperforms existing graph neural network and potential-based approaches for bilayer property prediction. 

\end{abstract}






\maketitle

\section{Introduction}
Stacked bilayer materials are formed by vertically assembling two monolayers. Although the layers are primarily coupled through weak van der Waals interactions, their properties can vary significantly with stacking configuration, interlayer distance, lattice mismatch, and twist angle. These stacking-dependent effects can give rise to electronic and optical properties that are absent in the corresponding isolated monolayers~\cite{li2010observation}. 
Density functional theory (DFT) with dispersion corrections, such as DFT-D3~\cite{grimme2010dftd3}, is commonly used to optimize bilayer structures and evaluate their properties. However, full DFT-D3 structural optimization is computationally expensive because a single monolayer pair can generate numerous stacking configurations, lateral translations, and interlayer distances. In existing bilayer material resources, 
the final bilayer structures are typically obtained through multi-stage optimization workflows involving interlayer-distance scans, in-plane registry optimization, and atomic relaxation before DFT property calculations are performed. The high computational cost of this process limits large-scale screening of bilayer materials and the construction of datasets for machine learning (ML).

In this work, we develop a complete workflow for stacked bilayer material property prediction, spanning bilayer structure construction, efficient structural optimization, and ML-based band gap prediction. 
We adopt Machine learning interatomic potentials (MLIPs), which learns accurate representations of interatomic interactions while achieving computational efficiencies, and employ MatterSim~\cite{deng2023mattersim} with DFT-D3 dispersion corrections~\cite{grimme2010dftd3} as an MLIP-D3 surrogate to replace costly DFT-D3 structural optimization. The resulting MatterSim-D3 workflow produces optimized bilayer structures at a fraction of the computational cost of conventional DFT-D3 optimization while maintaining comparable structural quality. 
To further exploit the unique interaction characteristics of bilayer materials, we propose BDIP-Net, an interaction-aware graph neural network that extends the potential-based modeling framework of PotNet \cite{lin2023potnet} to stacked bilayer systems. BDIP-Net constructs separate intra-layer and inter-layer edge sets, encodes them using Coulomb-based and London-dispersion-based potentials, and integrates the resulting messages through an interaction-message attention mechanism. By explicitly distinguishing strong intra-layer interactions from weak inter-layer van der Waals interactions, BDIP-Net learns interaction-specific representations while preserving the physics-informed advantages of potential-based message passing.

We evaluate our framework with  three computational stacked 2D materials databases, BiDB \cite{pakdel2024high} which covers homobilayers by optimizing multiple stacking configurations, HetDB which \cite{sauer2025dispersion} covers van der Waals heterobilayers, and SAMBA \cite{araujo2025high} that provides over 18,000 twisted homo- and hetero-bilayer structures. 
All three databases are built upon the  Computational 2D Materials Database (C2DB) \cite{haastrup2018computational} which is a comprehensive repository of first-principles-calculated monolayer materials.
Our experiments on these three benchmark bilayer datasets, BiDB, HetDB, and SAMBA, demonstrate that BDIP-Net consistently outperforms BiMat-ML, conventional bilayer-based baselines, PotNet, and SE-PotNet. Furthermore, MatterSim-D3 achieves nearly identical predictive performance to DFT-PBE-D3, validating the effectiveness of the proposed efficient optimization workflow. Together, these results show that combining efficient MLIP-based structural optimization with interaction-aware potential-based graph learning provides an effective and scalable framework for stacked bilayer material property prediction.

\section{Related Work}


Graph neural networks have become a widely adopted framework for materials property prediction by representing atoms as nodes and atomic interactions as edges~\cite{xie2018crystal,schutt2017schnet, chen2019graph, qiao2020orbnet}.  Most crystal GNNs, e.g., SchNet \cite{schutt2017schnet}, CGCNN \cite{xie2018crystal}, MEGNet \cite{chen2019graph}, and OrbNet \cite{qiao2020orbnet}, encode interatomic distances using radial basis function (RBF) expansions and learn interaction patterns through message passing.
Instead of directly encoding distances, PotNet ~\cite{lin2023potnet} transforms interatomic distances into potential values and uses the resulting potential-based features for message passing. Its experiment results show  physically informed interaction modeling can enhance crystal property prediction. However, PotNet was developed for general crystalline materials and treats all local atomic interactions using a unified representation. For stacked bilayer materials, the underlying interactions are inherently heterogeneous, consisting of strong intra-layer bonding and weak inter-layer van der Waals coupling. Consequently, directly applying PotNet to bilayer materials may not fully capture the distinct physical characteristics of these interaction types. Our proposed BDIP-Net builds upon the potential-based modeling philosophy of PotNet while explicitly distinguishing intra-layer and inter-layer interactions. By constructing separate interaction graphs and employing interaction-specific potential encodings together with interaction-message attention, BDIP-Net adapts physics-informed message passing to the unique interaction mechanisms of stacked bilayer materials.


Recent studies have explored machine learning approaches for predicting the properties of stacked two-dimensional (2D) bilayer materials. Among them, SE-PAINN~\cite{chen2024structural} explicitly distinguishes intra-layer and inter-layer interactions by constructing separate edge sets and processing them through dedicated message-passing branches. 
However, SE-PAINN relies on optimized bilayer structures generated through computationally intensive structural optimization workflows, limiting its scalability for large-scale screening. BiMat-ML~\cite{vuong2026bimatml} adopts a different strategy by predicting bilayer properties directly from monolayer structures, stacking configurations, and monolayer properties. By avoiding explicit bilayer structure construction and structural optimization, BiMat-ML significantly reduces computational cost. However, because the model operates on monolayer-level representations rather than reconstructed bilayer structures, it cannot directly learn atomic interactions arising from the stacked bilayer geometry.
In contrast, our work combines efficient MLIP-based structural optimization with interaction-aware graph learning on reconstructed bilayer structures, enabling both scalable structure generation and explicit modeling of intra-layer and inter-layer interactions.

\section{Method}

\begin{figure*}[t]
    \centering
    \includegraphics[width=1\linewidth]{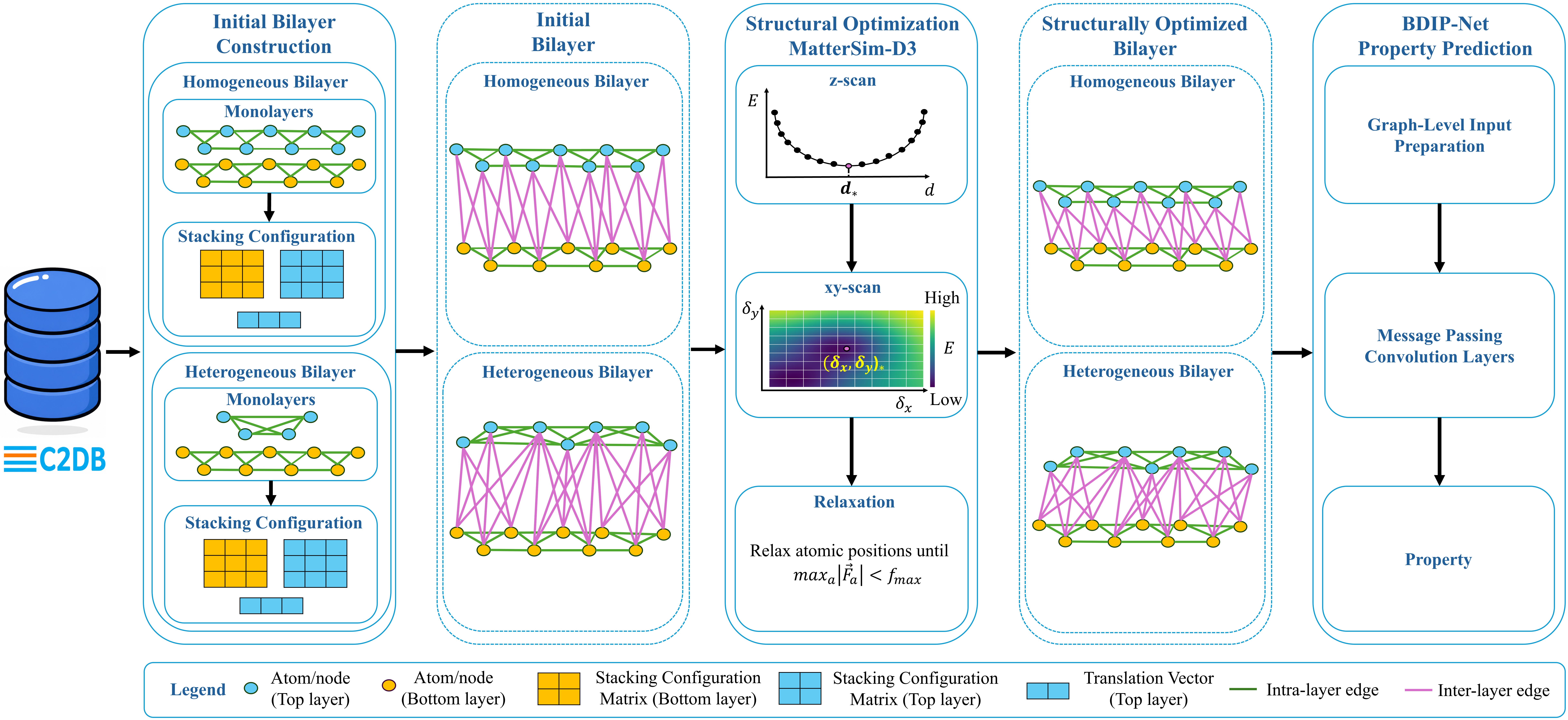}

    \caption{Workflow of the proposed bilayer material construction and property prediction framework. The workflow consists of two main stages. In the first stage, bilayer materials are constructed from monolayer structures obtained from C2DB. Two types of bilayers are considered, homogeneous bilayers formed from two identical monolayers and heterogeneous bilayers formed from two different monolayers. For each case, the initial bilayer construction uses a stacking configuration consisting of two \(3\times3\) stacking configuration matrices and a translation vector. The bottom and top stacking configuration matrices are applied to the corresponding layers to produce transformed layers, and the translation vector is applied to the top layer while keeping the bottom layer fixed to form the initial bilayer structure. The initial bilayer is then optimized using the MatterSim-D3 structural optimization workflow, including \(z\)-scan, \(xy\)-scan, and atomic relaxation, to obtain the structurally optimized bilayer. In the second stage, the structurally optimized bilayer material is used as input to BDIP-Net for graph-level input preparation, message passing convolution layers, and property prediction.}
    
    \label{fig:workflow}
\end{figure*}

The workflow of the proposed method is illustrated in Figure~\ref{fig:workflow}. It consists of two main stages, including bilayer material construction and bilayer property prediction. 

\subsection{MatterSim-D3-based Bilayer Structure Construction}
\label{subsec:bilayer_construction}

In general, the initial bilayer construction for BiDB, HetDB, and SAMBA starts from DFT-relaxed monolayer structures available in C2DB. Before stacking, each layer is transformed using a dataset-specific $3\times3$ stacking configuration matrix. This matrix defines the geometric operation applied to each layer before stacking, including enlarging the monolayer into a supercell, rotating the layer, and optionally flipping the layer along the out-of-plane direction. After these matrix-based transformations, a translation vector is applied to the top layer while keeping the bottom layer fixed. This vector includes optional in-plane translations along the \(x\) and \(y\) directions, and an out-of-plane translation along the \(z\) direction determined by the initial interlayer distance. We provide the background of stacked 2D materials representation in Appendix~\ref{app:background} and the detailed construction processes for BiDB, HetDB, and SAMBA in Appendix~\ref{app:bilayer_construction}.

In the original database workflows, the structural optimization of bilayer structures is performed using first-principles density functional theory (DFT) calculations with explicit treatments of van der Waals (vdW) interactions. All three workflows use DFT-PBE, based on the Perdew--Burke--Ernzerhof (PBE) exchange-correlation functional~\cite{perdew1996generalized}, but they employ different vdW treatments. BiDB and HetDB use the DFT-PBE-D3 framework, where the D3 dispersion correction is added to the PBE total energy to account for long-range vdW interactions~\cite{grimme2010dftd3}. The D3 term provides a computationally inexpensive correction for dispersion interactions. In contrast, SAMBA uses DFT-PBE with the optB86b vdW functional instead of the D3 correction, where optB86b is used to incorporate vdW interactions calculation~\cite{klimes2011van}. In these DFT-based workflows, the $z$-scan determines the optimal interlayer distance by minimizing the total energy, the $xy$-scan evaluates a $9\times9$ grid of in-plane translations to identify the lowest energy stacking registry when applied, and the final atomic relaxation optimizes the structure until the maximum atomic force is below the dataset-specific threshold $f_{\max}$. Although these procedures provide reliable optimized structures, each DFT-PBE total energy or maximum atomic force calculation is computationally expensive. Since the scan and relaxation steps require these calculations to be repeated many times for each candidate bilayer, the overall computational cost becomes substantial.

To reduce this cost, we adopt the MLIP-D3 structural optimization \cite{mishin2021machine}. Instead of repeatedly evaluating the DFT-PBE total energy and maximum atomic force, we use an MLIP for the PBE energy and force components and add the DFT-D3 dispersion correction to account for long-range vdW interactions. 
MLIP enables near-first-principles accuracy at computational costs that are orders of magnitude lower than DFT. This choice is computationally efficient because DFT-D3 has been classified as adding only a small additional computational cost, whereas vdW-DF methods such as optB86b-vdW increase the computational time by approximately \(50\%\) compared with a regular DFT-PBE calculation~\cite{klimes2011van,klimes2012perspective}. Evaluation results in \cite{sauer2025dispersion} show that using MatterSim~\cite{deng2023mattersim} together with the DFT-D3 correction, namely MatterSim-D3, gives the best overall performance among the evaluated MLIP-D3 models in reproducing bilayer structures optimized by DFT-PBE-D3, and DFT bandgap calculations based on bilayer structures optimized by MatterSim-D3 differ from those based on DFT-PBE-D3 optimized structures by only about \(0.02~\mathrm{eV}\) on average. Therefore, for all bilayer structures, including those from SAMBA, we use MatterSim-D3 to compute the total energy and maximum atomic force required in the \(z\)-scan, the optional \(xy\)-scan, and the relaxation. The detailed structural optimization process for each dataset is provided in Appendix~\ref{app:structural_optimization}. The resulting structurally optimized bilayer material \(M^{\mathrm{bi}}\) is saved in the Vienna Ab initio Simulation Package (VASP) format~\cite{kresse1996efficient} and then used as the input structure for the \mbox{BDIP-Net} model.

\subsection{BDIP-Net For Bilayer Property Prediction}

\begin{figure*}[ht]
    \centering
    \includegraphics[width=1\linewidth]{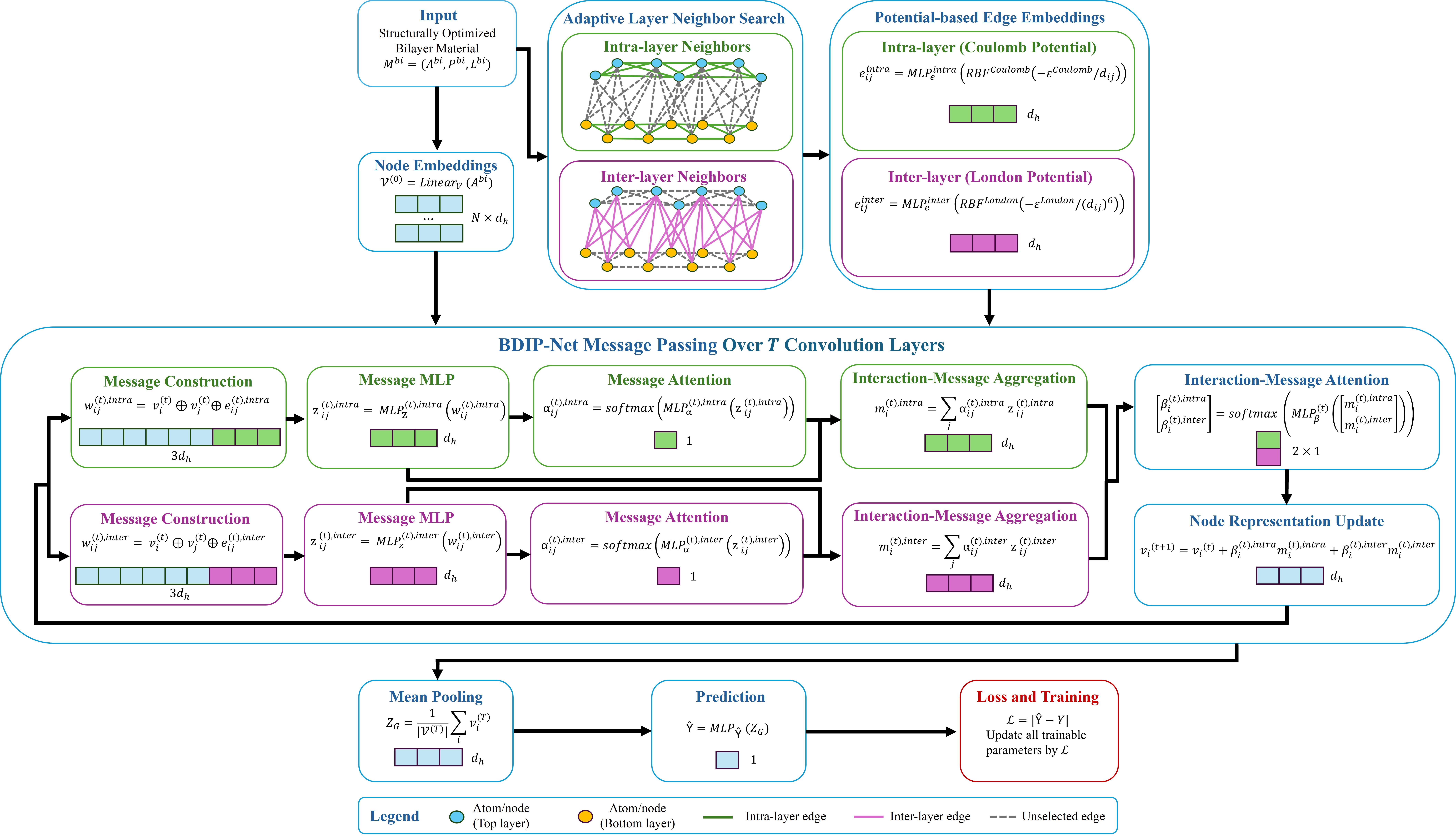}
     \caption{Architecture of the BDIP-Net model for stacked bilayer material property prediction. 
     The architecture consists of three main stages. In the first stage, the structurally optimized bilayer material is used to prepare graph-level inputs. Atomic features are transformed into initial node embeddings, and adaptive layer neighbor search is applied to construct intra-layer and inter-layer edge sets. Interaction-specific potential-based edge embeddings are then computed using a Coulomb-based potential for intra-layer edges and a London-based potential for inter-layer edges. In the second stage, BDIP-Net performs message passing over \(T\) convolution layers. In each layer, intra-layer and inter-layer messages are constructed and aggregated separately, and interaction-message attention is used to adaptively fuse the two interaction types when updating node representations. In the third stage, the final node representations are mean-pooled into a graph-level representation, which is passed to an output MLP to predict the target property. The model is trained end-to-end by minimizing the prediction loss.
     }
    \label{fig:BDIP-Net}
\end{figure*}

Figure \ref{fig:BDIP-Net} illusrates the overall training and prediction procedure of the proposed BDIP-Net model for stacked bilayer material property prediction. BDIP-Net consists of two main components. The first component prepares the graph-level inputs for message passing by finding two interaction-specific edge sets, namely intra-layer interaction edges and inter-layer interaction edges, initializing node embeddings, and computing interaction-specific potential-based edge embeddings. The second component performs node message passing, where intra-layer and inter-layer messages are aggregated separately and then adaptively fused to update node representations for property prediction.

\subsubsection{Graph-Level Input Preparation for Node Message Passing}
\label{sec:graph_edge_embedding}

Given a structurally optimized bilayer material $M^{\mathrm{bi}}=(A^{\mathrm{bi}},P^{\mathrm{bi}},L^{\mathrm{bi}})$, where $A^{\mathrm{bi}}\in\mathbb{R}^{N\times d_a}$ denotes the one-hot encoded atomic features of $N$ atoms, $P^{\mathrm{bi}}\in\mathbb{R}^{N\times3}$ denotes the atomic positions, and $L^{\mathrm{bi}}\in\mathbb{R}^{3\times3}$ denotes the lattice matrix, BDIP-Net first constructs interaction-specific edge sets for intra-layer and inter-layer interactions. Adaptive layer neighbor search is then applied separately to each interaction type under periodic boundary conditions. For each interaction type, the cutoff radius is adaptively increased until every atom has at least $N^{\mathrm{intra}}$ or $N^{\mathrm{inter}}$ valid neighbors after the corresponding filtering. This produces the intra-layer edge set $\mathcal{E}^{\mathrm{intra}}$ and the inter-layer edge set $\mathcal{E}^{\mathrm{inter}}$, preserving the distinction between the two interaction types for subsequent edge embedding and node message passing. Details of the adaptive layer neighbor search are provided in Appendix~\ref{app:layer_neighbor_search}.

After identifying the interaction-specific edge sets, BDIP-Net computes potential-based edge embeddings. This design is motivated by the PotNet ablation results~\cite{lin2023potnet}, which show that replacing raw distance-based features with physically motivated local potential features accounts for most of the performance improvement. Adding infinite potential-based features provides only a modest additional improvement over local potential features. BDIP-Net therefore adopts only local potential features, retaining the main benefit of the potential-based representation with a simpler design. Specifically, BDIP-Net follows PotNet by transforming the interatomic distance $d_{ij}$ of each intra-layer edge $(i,j)$ using a Coulomb-based potential. Since inter-layer interactions in stacked bilayer materials are typically dominated by weak van der Waals forces, they are commonly modeled using the Lennard--Jones potential~\cite{jones1924molecularfieldsII}. BDIP-Net instead adopts a London-dispersion-based potential~\cite{wagner2015london} as a simpler formulation for capturing the dominant attractive $d^{-6}$ component of these interactions. The scalar potentials are
$V_{\mathrm{Coulomb}}(i,j)=-\epsilon^{\mathrm{Coulomb}}/d_{ij}$ and
$V_{\mathrm{London}}(i,j)=-\epsilon^{\mathrm{London}}/(d_{ij})^{6}$,
where $\epsilon^{\mathrm{Coulomb}}$ and
$\epsilon^{\mathrm{London}}$ are scaling coefficients.
\begin{equation}
\begin{aligned}
e_{ij}^{\mathrm{intra}}
&=
\mathrm{MLP}_{e}^{\mathrm{intra}}
\!\left(
\mathrm{RBF}^{\mathrm{Coulomb}}
\!\left(
-\epsilon^{\mathrm{Coulomb}}/d_{ij}
\right)
\right)
\in\mathbb{R}^{d_h},\\
e_{ij}^{\mathrm{inter}}
&=
\mathrm{MLP}_{e}^{\mathrm{inter}}
\!\left(
\mathrm{RBF}^{\mathrm{London}}
\!\left(
-\epsilon^{\mathrm{London}}/(d_{ij})^{6}
\right)
\right)
\in\mathbb{R}^{d_h},
\end{aligned}
\end{equation}

where $\mathrm{RBF}^{\mathrm{Coulomb}}$ and $\mathrm{RBF}^{\mathrm{London}}$ denote separate RBF expansions, while $\mathrm{MLP}_{e}^{\mathrm{intra}}$ and $\mathrm{MLP}_{e}^{\mathrm{inter}}$ are interaction-specific edge networks. Details of the RBF expansions are provided in Appendix~\ref{app:rbf}. Next, the atomic node embeddings are initialized as
$\mathcal{V}^{(0)}=\mathrm{Linear}_{\mathcal{V}}(A^{\mathrm{bi}})
\in\mathbb{R}^{N\times d_h}$,
where $\mathrm{Linear}_{\mathcal{V}}$ denotes a linear projection.

\subsubsection{Message Passing and Property Prediction}
\label{sec:message_passing_prediction}

This subsection describes the second component of BDIP-Net, which updates node representations through message passing and then predicts the target bilayer material property. Specifically, BDIP-Net applies $T$ message-passing convolution layers to update the initial node embeddings obtained from Section~\ref{sec:graph_edge_embedding}. At each layer, the representation of each node is updated by aggregating information from two interaction-specific branches, namely an intra-layer branch and an inter-layer branch. The intra-layer branch aggregates edge messages from neighboring nodes within the same layer, while the inter-layer branch aggregates edge messages from neighboring nodes across different layers. Within each branch, each node receives edge messages from its neighboring nodes, and BDIP-Net computes a scalar attention weight for each edge message to measure its importance during interaction-message aggregation. The weighted edge messages in each branch are then aggregated to produce one intra-layer interaction-message and one inter-layer interaction message for each node. These two interaction-messages are further adaptively fused by interaction-message attention, which computes scalar attention weights to learn the relative contribution of the intra-layer and inter-layer interaction messages when updating each node representation. 
The use of scalar attention weights for both edge messages and interaction-specific messages is conceptually inspired by the slot-based attention idea in SlotGAT~\cite{zhou2023slotgat}, where separate representation slots are maintained and their relative importance is learned by attention. However, unlike SlotGAT, where slots correspond to node types in heterogeneous graphs, BDIP-Net applies scalar attention weights to edge messages within each interaction branch and to interaction-specific messages across the intra-layer and inter-layer branches. For each interaction type
$\eta \in \{\mathrm{intra},\mathrm{inter}\}$ and each edge
$(i,j) \in \mathcal{E}^{\eta}$, BDIP-Net constructs the pairwise interaction representation as
\begin{equation}
w_{ij}^{(t),\eta}
=
v_i^{(t)} \oplus v_j^{(t)} \oplus e_{ij}^{\eta}
\in \mathbb{R}^{3d_h},
\label{eq:interaction_rep}
\end{equation}
where $\oplus$ denotes feature concatenation and
$e_{ij}^{\eta}$ denotes the edge feature associated with interaction type $\eta$.
The pairwise interaction representation is then transformed into an interaction-specific edge message:
\begin{equation}
z_{ij}^{(t),\eta}
=
\mathrm{MLP}_{z}^{(t),\eta}
\left(
w_{ij}^{(t),\eta}
\right)
\in \mathbb{R}^{d_h},
\label{eq:edge_message}
\end{equation}
where $\mathrm{MLP}_{z}^{(t),\eta}$ maps the
$3d_h$-dimensional pairwise interaction representation to a
$d_h$-dimensional edge message for interaction type $\eta$.
The attention weight of each edge is computed from the transformed edge message:
\begin{equation}
\alpha_{ij}^{(t),\eta}
=
\operatorname{softmax}_{j \in \mathcal{N}^{\eta}_{(i)}}
\left(
\mathrm{MLP}_{\alpha}^{(t),\eta}
\left(
z_{ij}^{(t),\eta}
\right)
\right)
\in \mathbb{R},
\label{eq:edge_attention}
\end{equation}
where $\mathrm{MLP}_{\alpha}^{(t),\eta}$ maps each
$d_h$-dimensional edge message to an attention score, and the softmax is normalized over the neighbors of node $i$ associated with interaction type $\eta$.
The interaction-specific message of node $i$ is then aggregated as
\begin{equation}
m_i^{(t),\eta}
=
\sum_{j \in \mathcal{N}^{\eta}_{(i)}}
\alpha_{ij}^{(t),\eta}
z_{ij}^{(t),\eta}
\in \mathbb{R}^{d_h}.
\label{eq:interaction_message_update}
\end{equation}

After the intra-layer and inter-layer messages are obtained, BDIP-Net applies interaction-message attention fusion to adaptively combine them for each atom. The interaction-message attention weights are computed as
\begin{equation}
\begin{bmatrix}
\beta_i^{(t),\mathrm{intra}}\\
\beta_i^{(t),\mathrm{inter}}
\end{bmatrix}
=
\operatorname{softmax}
\left(
\mathrm{MLP}_{\beta}^{(t)}
\left(
\begin{bmatrix}
m_i^{(t),\mathrm{intra}}\\
m_i^{(t),\mathrm{inter}}
\end{bmatrix}
\right)
\right)
\in \mathbb{R}^{2},
\label{eq:slot_attention}
\end{equation}
where $\mathrm{MLP}_{\beta}^{(t)}$ is shared across the intra-layer and inter-layer interaction-messages at layer $t$. The MLP maps the two $d_h$-dimensional interaction-messages to two attention scores, and the softmax normalizes the scores over the two message slots. The node representation is then updated by adding the weighted interaction messages to the previous-layer representation:
\begin{equation}
v_i^{(t+1)}
=
v_i^{(t)}
+
\beta_i^{(t),\mathrm{intra}}
m_i^{(t),\mathrm{intra}}
+
\beta_i^{(t),\mathrm{inter}}
m_i^{(t),\mathrm{inter}}
\in \mathbb{R}^{d_h}.
\label{eq:node_update}
\end{equation}

After $T$ message-passing layers, the final node representations $\{v_i^{(T)}\}_{i\in\mathcal{V}}$ are averaged to obtain the graph-level representation:
\begin{equation}
Z_{\mathcal{G}}
=
\frac{1}{|\mathcal{V}|}
\sum_{i\in\mathcal{V}} v_i^{(T)}
\in \mathbb{R}^{d_h}.
\label{eq:graph_representation}
\end{equation}

The graph-level representation $Z_{\mathcal{G}}$ is passed through an output MLP to predict the target bilayer property as
$\hat{Y}=\mathrm{MLP}_{\hat{Y}}(Z_{\mathcal{G}})$. The entire framework is trained end-to-end by minimizing the mean absolute error loss $\mathcal{L}=|\hat{Y}-Y|$. The output $\mathrm{MLP}_{\hat{Y}}$ uses a Linear--ShiftedSoftplus--Linear network architecture, while all other MLPs in BDIP-Net use a Linear--SiLU--Linear network architecture, where ShiftedSoftplus and SiLU denote the activation functions between the two linear layers. MLPs with different subscripts or superscripts, including those for different interaction types and different convolution layers, are independently parameterized. The pseudocode of our BDIP-Net is shown in Algorithm \ref{alg:BDIP-Net} of Appendix \ref{app:BDIP-Net}.

\section{Experiments}

\subsection{Experiment Settings}
\label{sec:exp_setting}
\begin{figure}[H]
    \centering
    \includegraphics[width=1\linewidth]{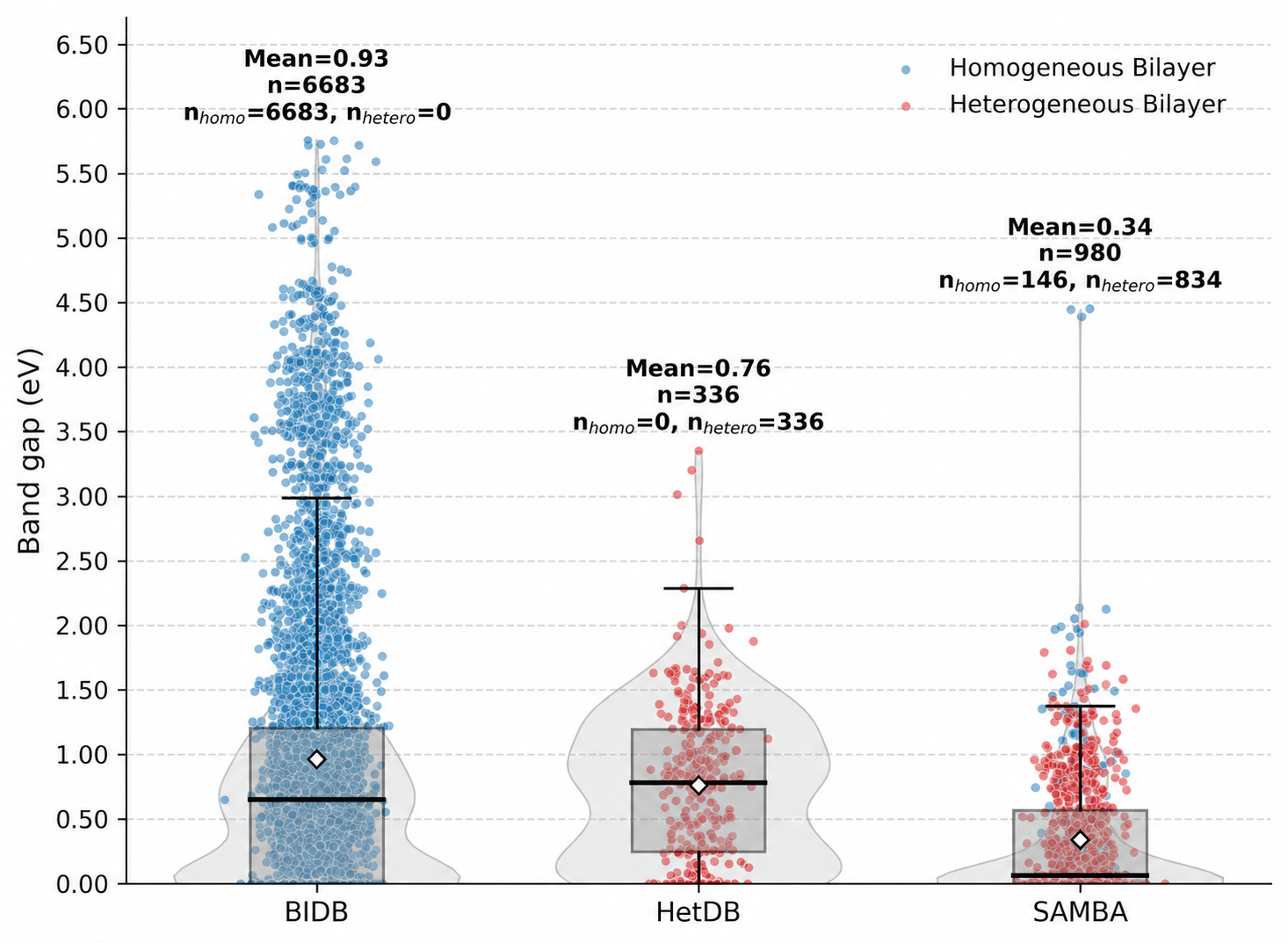}
    \caption{Bandgap distribution in BiDB, HetDB, and SAMBA.}
    \label{fig:band_gap_box_plot}
\end{figure}

\noindent\textbf{Datasets.} We follow Bimat-ML~\cite{vuong2026bimatml} and use the same processed BiDB and HetDB datasets. The SAMBA dataset contains more than 18,000 twisted bilayer structures generated from 63 monolayers in the C2DB database. We use 980 bilayers (144 homobilayers and 836 heterobilayers) with available DFT-optimized structures and bandgap values. For all three datasets, we use bandgap as the target property. Fig.~\ref{fig:band_gap_box_plot} shows the composition and bandgap distributions of the three datasets. BiDB contains 6,683 homobilayers with a mean bandgap of \(0.93\) eV, while HetDB contains 336 heterobilayers with a mean bandgap of \(0.76\) eV. SAMBA contains 146 homobilayers and 834 heterobilayers, with mean bandgaps of \(0.66\) and \(0.28\) eV, respectively. The three datasets provide complementary coverage of aligned homobilayers, heterobilayers, and twisted bilayer systems.  The differences in dataset size, bilayer type, bandgap distribution, and bilayer-monolayer bandgap relationships form an ideal benchmark for evaluation. 


\noindent\textbf{Cross-validation strategy.}
We perform 4-fold cross-validation by splitting each dataset at the monolayer-pair level rather than the bilayer level. All bilayers constructed from the same monolayer pair are assigned to the same fold. For homogeneous bilayers, the pair is represented as \((M,M)\), whereas for heterogeneous bilayers it is represented as \((M_A,M_B)\). This strategy prevents the same monolayer pair from appearing in both the training and test sets, thereby evaluating the models' ability to generalize to bilayers constructed from unseen monolayer pairs.


\noindent\textbf{BDIP-Net.}
Our BDIP-Net uses the MatterSim-D3 structural optimization procedure described in Section~\ref{subsec:bilayer_construction}. 
We also report the results from DFT-PBE-D3  where the optimized bilayer structures provided by BiDB, HetDB, and SAMBA are directly converted into the VASP format and used as model inputs. For BiDB, HetDB, and SAMBA, we use a consistent BDIP-Net hyperparameter configuration, with only the batch size adjusted for HetDB. The model consists of three convolutional layers (\(T=3\)) and uses adaptive layer neighbor search with \(N^{\mathrm{intra}}=16\) and \(N^{\mathrm{inter}}=16\), starting from initial search radii of \(R^{\mathrm{intra}}=R^{\mathrm{inter}}=4\). Both potential scaling factors are set to \(\epsilon^{\mathrm{Coulomb}}=\epsilon^{\mathrm{London}}=1\), and the hidden dimension is set to \(d_h=128\) for both node and edge embeddings. The model is trained for 500 epochs using the Adam optimizer with a learning rate of \(2\times10^{-4}\). The batch size is set to 128 for BiDB and SAMBA and 16 for HetDB.

\noindent\textbf{Baselines.}
We compare with Bimat-ML~\cite{vuong2026bimatml}, SE-CGCNN~\cite{chen2024structural}, SE-MEGNET~\cite{chen2024structural}, SE-PAINN~\cite{chen2024structural} and PotNet \cite{lin2023potnet}. BDIP-Net and the bilayer-based baselines are evaluated under the two structural optimization settings described above, as indicated in the Optimization column of the result tables. In our experiments, Bimat-ML uses only monolayer structures and stacking configuration information as inputs, without using monolayer properties, ensuring a fair comparison with the other models. Since Bimat-ML does not take an optimized bilayer structure as input, its entry in the Optimization column is marked as N/A. 
All experiments in this work are conducted on an NVIDIA H200 GPU with 140~GB of memory.

\subsection{Experiment Results}

Tables~\ref{tab:combined_bandgap_results} summarizes the single-domain bandgap prediction results on BiDB, HetDB, and SAMBA, respectively. Each table reports the MAE, MSE, RMSE, and \(R^2\) of BDIP-Net and the baseline models using bilayer structures optimized by DFT-PBE-D3 and MatterSim-D3. Bimat-ML, which does not use optimized bilayer structures or monolayer property inputs, is included as a structure-free baseline. 

\begin{table}[t]
\centering
\caption{Regression performance on the BiDB, HetDB, and SAMBA datasets for bandgap prediction.}
\label{tab:combined_bandgap_results}

\resizebox{\columnwidth}{!}{%
\begin{tabular}{lllcccc}
\toprule
\textbf{Dataset} &
\textbf{Model} &
\textbf{Optimization} &
\textbf{MAE} $\downarrow$ &
\textbf{MSE} $\downarrow$ &
\textbf{RMSE} $\downarrow$ &
\textbf{R$^2$} $\uparrow$ \\
\midrule

\multirow{11}{*}{BiDB}
& Bimat-ML
& N/A
& $0.35 \pm 0.05$
& $0.35 \pm 0.08$
& $0.59 \pm 0.06$
& $0.68 \pm 0.06$ \\

& \multirow{2}{*}{SE-CGCNN}
& DFT-PBE-D3
& $0.38 \pm 0.05$
& $0.49 \pm 0.19$
& $0.69 \pm 0.09$
& $0.68 \pm 0.06$ \\

&
& MatterSim-D3
& $0.38 \pm 0.05$
& $0.50 \pm 0.18$
& $0.70 \pm 0.13$
& $0.66 \pm 0.08$ \\

& \multirow{2}{*}{SE-MEGNET}
& DFT-PBE-D3
& $0.36 \pm 0.05$
& $0.37 \pm 0.13$
& $0.60 \pm 0.11$
& $0.67 \pm 0.06$ \\

&
& MatterSim-D3
& $0.36 \pm 0.06$
& $0.37 \pm 0.12$
& $0.60 \pm 0.10$
& $0.67 \pm 0.08$ \\

& \multirow{2}{*}{SE-PAINN}
& DFT-PBE-D3
& $0.36 \pm 0.04$
& $0.40 \pm 0.08$
& $0.63 \pm 0.07$
& $0.64 \pm 0.06$ \\

&
& MatterSim-D3
& $0.36 \pm 0.06$
& $0.40 \pm 0.07$
& $0.63 \pm 0.06$
& $0.63 \pm 0.07$ \\

& \multirow{2}{*}{PotNet}
& DFT-PBE-D3
& $0.33 \pm 0.02$
& $0.36 \pm 0.06$
& $0.60 \pm 0.06$
& $0.66 \pm 0.03$ \\

&
& MatterSim-D3
& $0.33 \pm 0.03$
& $0.36 \pm 0.07$
& $0.60 \pm 0.06$
& $0.65 \pm 0.04$ \\

& \multirow{2}{*}{BDIP-Net}
& DFT-PBE-D3
& ${0.30 \pm 0.04}$
& ${0.32 \pm 0.10}$
& $0.57 \pm 0.10$
& $0.72 \pm 0.05$ \\

&
& MatterSim-D3
& $\mathbf{0.30 \pm 0.03}$
& $\mathbf{0.32 \pm 0.10}$
& $\mathbf{0.56 \pm 0.09}$
& $\mathbf{0.73 \pm 0.05}$ \\
\midrule
\multirow{11}{*}{HetDB}
& Bimat-ML
& N/A
& $0.16 \pm 0.01$
& $0.09 \pm 0.04$
& $0.29 \pm 0.06$
& $0.76 \pm 0.12$ \\

& \multirow{2}{*}{SE-CGCNN}
& DFT-PBE-D3
& $0.12 \pm 0.02$
& $0.03 \pm 0.02$
& $0.18 \pm 0.02$
& $0.91 \pm 0.02$ \\

&
& MatterSim-D3
& $0.12 \pm 0.01$
& $0.03 \pm 0.02$
& $0.17 \pm 0.04$
& $0.92 \pm 0.05$ \\

& \multirow{2}{*}{SE-MEGNET}
& DFT-PBE-D3
& $0.19 \pm 0.02$
& $0.09 \pm 0.03$
& $0.29 \pm 0.06$
& $0.76 \pm 0.08$ \\

&
& MatterSim-D3
& $0.19 \pm 0.02$
& $0.09 \pm 0.04$
& $0.30 \pm 0.06$
& $0.75 \pm 0.07$ \\

& \multirow{2}{*}{SE-PAINN}
& DFT-PBE-D3
& $0.11 \pm 0.02$
& $0.03 \pm 0.02$
& $0.17 \pm 0.02$
& $0.91 \pm 0.02$ \\

&
& MatterSim-D3
& $0.11 \pm 0.01$
& $0.03 \pm 0.02$
& $0.17 \pm 0.05$
& $0.90 \pm 0.03$ \\

& \multirow{2}{*}{PotNet}
& DFT-PBE-D3
& $0.11 \pm 0.02$
& $0.04 \pm 0.03$
& $0.21 \pm 0.06$
& $0.89 \pm 0.07$ \\

&
& MatterSim-D3
& $0.11 \pm 0.02$
& $0.05 \pm 0.02$
& $0.21 \pm 0.05$
& $0.89 \pm 0.05$ \\

& \multirow{2}{*}{BDIP-Net}
& DFT-PBE-D3
& ${0.09} \pm 0.02$
& ${0.02} \pm 0.02$
& ${0.14} \pm 0.05$
& ${0.94} \pm 0.05$ \\

&
& MatterSim-D3
& $\mathbf{0.09 \pm 0.01}$
& $\mathbf{0.02 \pm 0.02}$
& $\mathbf{0.15 \pm 0.06}$
& $\mathbf{0.93 \pm 0.05}$ \\
\midrule
\multirow{11}{*}{SAMBA}
& Bimat-ML
& N/A
& $0.18 \pm 0.03$
& $0.13 \pm 0.09$
& $0.35 \pm 0.12$
& $0.66 \pm 0.12$ \\

& \multirow{2}{*}{SE-CGCNN}
& DFT-PBE-D3
& $0.16 \pm 0.02$
& $0.09 \pm 0.08$
& $0.29 \pm 0.11$
& $0.71 \pm 0.11$ \\

&
& MatterSim-D3
& $0.16 \pm 0.02$
& $0.09 \pm 0.06$
& $0.29 \pm 0.10$
& $0.70 \pm 0.09$ \\

& \multirow{2}{*}{SE-MEGNET}
& DFT-PBE-D3
& $0.15 \pm 0.02$
& $0.08 \pm 0.05$
& $0.28 \pm 0.08$
& $0.73 \pm 0.04$ \\

&
& MatterSim-D3
& $0.15 \pm 0.02$
& $0.08 \pm 0.05$
& $0.28 \pm 0.08$
& $0.73 \pm 0.06$ \\

& \multirow{2}{*}{SE-PAINN}
& DFT-PBE-D3
& $0.15 \pm 0.02$
& $0.09 \pm 0.07$
& $0.27 \pm 0.10$
& $0.73 \pm 0.09$ \\

&
& MatterSim-D3
& $0.15 \pm 0.02$
& $0.09 \pm 0.06$
& $0.28 \pm 0.11$
& $0.73 \pm 0.09$ \\

& \multirow{2}{*}{PotNet}
& DFT-PBE-D3
& $0.13 \pm 0.02$
& $0.07 \pm 0.09$
& $0.26 \pm 0.10$
& $0.74 \pm 0.09$ \\

&
& MatterSim-D3
& $0.13 \pm 0.02$
& $0.09 \pm 0.06$
& $0.26 \pm 0.10$
& $0.73 \pm 0.08$ \\

& \multirow{2}{*}{BDIP-Net}
& DFT-PBE-D3
& ${0.10} \pm 0.02$
& $0.05 \pm 0.04$
& $0.21 \pm 0.07$
& ${0.83} \pm 0.03$ \\

&
& MatterSim-D3
& $\mathbf{0.10 \pm 0.01}$
& $\mathbf{0.04 \pm 0.02}$
& $\mathbf{0.20 \pm 0.05}$
& $\mathbf{0.83 \pm 0.03}$ \\

\bottomrule
\end{tabular}%
}
\end{table}

\label{app:sig_test}

\begin{table}[t]
\centering
\caption{P-values of pairwise comparisons against BDIP-Net over combined 3 datasets.}
\label{tab:p-value}
\begin{tabular}{lcccc}
\toprule
\textbf{Model} &
\textbf{MAE} &
\textbf{MSE} &
\textbf{RMSE} &
\textbf{R$^2$} \\
\midrule
Bimat-ML
& 0.000 & 0.000 & 0.000 & 0.000 \\

SE-CGCNN
& 0.000 & 0.006 & 0.000 & 0.001 \\

SE-MEGNET
& 0.000 & 0.000 & 0.000 & 0.000 \\

SE-PAINN
& 0.000 & 0.001 & 0.001 & 0.000 \\

PotNet
& 0.000 & 0.001 & 0.000 & 0.000 \\

BDIP-Net$^{-}$
& 0.002 & 0.002 & 0.002 & 0.001 \\
\bottomrule
\end{tabular}
\end{table}

\subsubsection{Prediction Performance}

As shown in Table~\ref{tab:combined_bandgap_results}, our BDIP-Net achieves the best  performance across all four metrics (MAE, MSE, RMSE, and \(R^2\)) and and  all three datasets (BiDB, HetDB, and SAMBA). Specifically, compared with the best-performed baseline PotNet,  BDIP-Net increases the \(R^2\) from \(0.65\) to \(0.73\) on BiDB, from \(0.89\) to \(0.93\) on HetDB, from \(0.73\) to \(0.83\) on SAMBA. 

\noindent\textbf{Statistical significance.}
We perform one-sided paired \(t\)-tests to compare BDIP-Net with each baseline model for MAE, MSE, RMSE, and \(R^2\), with the alternative hypothesis that BDIP-Net achieves lower MAE, MSE, and RMSE and a higher \(R^2\). For each comparison, the four-fold results from BiDB, HetDB, and SAMBA are combined to form 12 paired observations. All comparisons between BDIP-Net and the baseline models yield \(p<0.01\), as shown in 
Table~\ref{tab:p-value}, 
indicating that the improvements of BDIP-Net are statistically significant. 

\subsubsection{Structural Optimization and Its Effect on Prediction}


Table~\ref{tab:combined_bandgap_results} also reports the bandgap prediction results obtained using DFT-PBE-D3-optimized structure. Both BDIP-Net and baseline models  exhibit only minor variations when DFT-PBE-D3-optimized structures are replaced with MatterSim-D3-optimized structures. Overall, the small differences between the two optimization settings indicate that MatterSim-D3 preserves the structural information required for bandgap prediction and achieves performance comparable to that obtained using DFT-PBE-D3-optimized structures while significantly reducing the computational cost.


\subsubsection{Ablation Study}
\label{sec:ablation_study}

We conduct an ablation study on the message-passing operation and interaction-message attention module of BDIP-Net by comparing the original model with an ablated variant, denoted as BDIP-Net$^{-}$. In the intra-layer and inter-layer branches of BDIP-Net$^{-}$, the original message-passing operation, which assigns a single scalar softmax attention weight to each neighboring edge message, is replaced with the PotNet-style message-passing operation based on a vector of feature-wise sigmoid gates. In addition, the scalar softmax interaction-message attention module, which assigns a scalar attention weight to each aggregated intra-layer and inter-layer message, is removed, and the two messages are combined by direct summation. This comparison evaluates the overall effect of replacing the proposed message-passing operation and scalar softmax interaction-message attention module with PotNet-style message passing and direct message summation. 

\begin{table}[t]
\centering
\caption{Ablation study of BDIP-Net$^{-}$ and BDIP-Net using MatterSim-D3-optimized structures.}
\label{tab:ablation_study}
\resizebox{\columnwidth}{!}{%
\begin{tabular}{llcccc}
\toprule
\textbf{Dataset} &
\textbf{Model} &
\textbf{MAE} $\downarrow$ &
\textbf{MSE} $\downarrow$ &
\textbf{RMSE} $\downarrow$ &
\textbf{R$^2$} $\uparrow$ \\
\midrule

\multirow{2}{*}{BiDB}
& BDIP-Net$^{-}$
& $0.32 \pm 0.03$
& $0.34 \pm 0.12$
& $0.59 \pm 0.11$
& $0.69 \pm 0.04$ \\

& \textbf{BDIP-Net}
& $\mathbf{0.30 \pm 0.03}$
& $\mathbf{0.32 \pm 0.11}$
& $\mathbf{0.56 \pm 0.10}$
& $\mathbf{0.73 \pm 0.05}$ \\
\midrule
\multirow{2}{*}{HetDB}
& BDIP-Net$^{-}$
& $0.10 \pm 0.01$
& $0.03 \pm 0.02$
& $0.18 \pm 0.06$
& $0.91 \pm 0.07$ \\

& \textbf{BDIP-Net}
& $\mathbf{0.09 \pm 0.01}$
& $\mathbf{0.02 \pm 0.02}$
& $\mathbf{0.15 \pm 0.06}$
& $\mathbf{0.93 \pm 0.05}$ \\
\midrule
\multirow{2}{*}{SAMBA}
& BDIP-Net$^{-}$
& $0.13 \pm 0.04$
& $0.08 \pm 0.06$
& $0.26 \pm 0.11$
& $0.75 \pm 0.11$ \\

& \textbf{BDIP-Net}
& $\mathbf{0.10 \pm 0.01}$
& $\mathbf{0.04 \pm 0.02}$
& $\mathbf{0.20 \pm 0.05}$
& $\mathbf{0.83 \pm 0.03}$ \\

\bottomrule
\end{tabular}%
}
\end{table}

Table~\ref{tab:ablation_study} shows that BDIP-Net consistently outperforms BDIP-Net$^{-}$ across all three datasets. Specifically,  the  \(R^2\) values increase from \(0.69\) to \(0.73\) on BiDB, from \(0.91\) to \(0.93\) on HetDB, and from \(0.75\) to \(0.83\) on SAMBA, respectively. Similar to the comparisons with the other baseline models, one-sided paired \(t\)-tests on the combined four-fold results from BiDB, HetDB, and SAMBA show that the improvements of BDIP-Net over BDIP-Net\(^{-}\) are statistically significant across all evaluation metrics (\(p<0.01\)).

\subsubsection{Runtime Analysis and Computational Efficiency}

\noindent\textbf{Structural Optimization Runtime.}
BDIP-Net adopts the efficient MatterSim-D3 to performs explicit structural optimization rather than using expensive DFT-PBE-D3-based energy and force calculations. The complete structural optimization takes an average of 4.5 seconds per material on BiDB, 3.5 seconds on HetDB, and 13.5 seconds on SAMBA. The runtime differences among the three datasets mainly arise from their different optimization workflows and relaxation criteria. 
The structural optimization time is longest for SAMBA because its workflow additionally includes full \(z\)-scan, \(xy\)-scan, and relaxation procedures for twisted bilayers. 
In contrast, obtaining a \mbox{DFT-PBE-D3}-optimized structure typically requires hours of DFT-based structural optimization. 

\noindent\textbf{Training and Inference Runtime.}
On BiDB, \mbox{BDIP-Net} requires \(575~\mathrm{s}\) for training,  comparable to PotNet (\(550~\mathrm{s}\)) and SE-MEGNET (\(475~\mathrm{s}\)), but longer than SE-CGCNN (\(400~\mathrm{s}\)), SE-PAINN (\(300~\mathrm{s}\)), and BiMat-ML (\(285~\mathrm{s}\)). On HetDB, the training time of \mbox{BDIP-Net} is \(40~\mathrm{s}\), compared with \(35~\mathrm{s}\) for PotNet, \(25~\mathrm{s}\) for SE-MEGNET, \(20~\mathrm{s}\) for SE-CGCNN, and \(12~\mathrm{s}\) for both SE-PAINN and BiMat-ML. On SAMBA, \mbox{BDIP-Net} requires \(80~\mathrm{s}\) while PotNet, SE-MEGNET, SE-CGCNN, SE-PAINN, and BiMat-ML requires \(70~\mathrm{s}\), \(65~\mathrm{s}\), \(60~\mathrm{s}\), \(50~\mathrm{s}\), and \(48~\mathrm{s}\), respectively. Despite its modestly higher training cost, BDIP-Net achieves the best prediction accuracy across all three datasets. All methods exhibit similar inference efficiency, with an average per-sample inference time of approximately 0.05 s. Detailed training runtime comparisons are provided in Appendix \ref{app:runningtime} (Figure~\ref{fig:training_time}). 


\subsubsection{Cross-dataset Domain Generalization}


We use HetDB and a subset of BiDB (denoted as BiDB$^{*}$) as the source-domain training datasets, and the full SAMBA  as the target-domain test set. BiDB contains 6,683 homobilayers, while HetDB contains 336 heterobilayers. The SAMBA test set contains 980 bilayers, including 146 homobilayers and 834 heterobilayers. Directly combining the full BiDB dataset with HetDB would make the training data heavily dominated by homobilayers. BiDB$^{*}$  contains 249 bilayers formed from 39 monolayers overlapping with SAMBA. BiDB and HetDB have 39 and 13 monolayers overlapping with SAMBA, respectively, and the complete lists are provided in the Appendix~\ref{app:cross_domain_mono_overlap}. 
Since PotNet achieves the best overall single-domain performance among the baseline models
, it is selected as the representative baseline for comparison.
To evaluate how prior exposure to constituent monolayers affects cross-dataset generalization, we partition the SAMBA test set based on monolayer overlap with the training data, irrespective of layer ordering. The test set comprises 147 bilayers with no monolayer overlap, 386 with one-monolayer overlap, and 447 with two-monolayer overlap. To further distinguish previously observed bilayer compositions from novel combinations of known monolayers, the two-monolayer-overlap subset is divided into 121 exact-pair-overlap bilayers, in which the same monolayer pair appears in the training set (regardless of layer ordering), and 326 bilayers containing known monolayers but unseen pairings.

\begin{table}[t]
\centering
\caption{BiDB$^{*}$+HetDB-to-SAMBA domain generalization performance.}
\label{tab:cross_domain_generalization_bidb_star_hetdb_samba}
\renewcommand{\arraystretch}{1.10}

\resizebox{\columnwidth}{!}{%
\begin{tabular}{lclcccc}
\toprule
\textbf{Setting} &
\textbf{\makecell{Number of\\Bilayers}} &
\textbf{Model} &
\textbf{MAE} $\downarrow$ &
\textbf{MSE} $\downarrow$ &
\textbf{RMSE} $\downarrow$ &
\textbf{R$^2$} $\uparrow$ \\
\midrule

\multirow{2}{*}{Full SAMBA}
& \multirow{2}{*}{980}
& BDIP-Net
& \textbf{0.27}
& \textbf{0.16}
& \textbf{0.40}
& \textbf{0.40} \\
&
& PotNet
& 0.34
& 0.22
& 0.47
& 0.15 \\

\midrule

\multirow{2}{*}{No-monolayer-overlap}
& \multirow{2}{*}{147}
& BDIP-Net
& \textbf{0.43}
& 0.38
& 0.62
& -0.62 \\
&
& PotNet
& 0.48
& \textbf{0.34}
& \textbf{0.59}
& \textbf{-0.47} \\

\midrule

\multirow{2}{*}{One-monolayer-overlap}
& \multirow{2}{*}{386}
& BDIP-Net
& \textbf{0.31}
& \textbf{0.18}
& \textbf{0.43}
& \textbf{-0.35} \\
&
& PotNet
& 0.41
& 0.29
& 0.54
& -1.13 \\

\midrule

\multirow{2}{*}{Two-monolayer-overlap}
& \multirow{2}{*}{447}
& BDIP-Net
& \textbf{0.17}
& \textbf{0.06}
& \textbf{0.25}
& \textbf{0.82} \\
&
& PotNet
& 0.23
& 0.12
& 0.35
& 0.67 \\

\midrule

\multirow{2}{*}{Exact-pair-overlap}
& \multirow{2}{*}{121}
& BDIP-Net
& \textbf{0.09}
& \textbf{0.02}
& \textbf{0.13}
& \textbf{0.98} \\
&
& PotNet
& \textbf{0.09}
& \textbf{0.02}
& 0.14
& 0.97 \\

\midrule

\multirow{2}{*}{%
\makecell[l]{Two-monolayer-overlap excluding\\exact-pair-overlap}}
& \multirow{2}{*}{326}
& BDIP-Net
& \textbf{0.20}
& \textbf{0.08}
& \textbf{0.29}
& \textbf{0.54} \\
&
& PotNet
& 0.29
& 0.16
& 0.40
& 0.12 \\

\bottomrule
\end{tabular}%
}
\end{table}

\begin{figure}[t]
    \centering
    \includegraphics[width=1\linewidth]{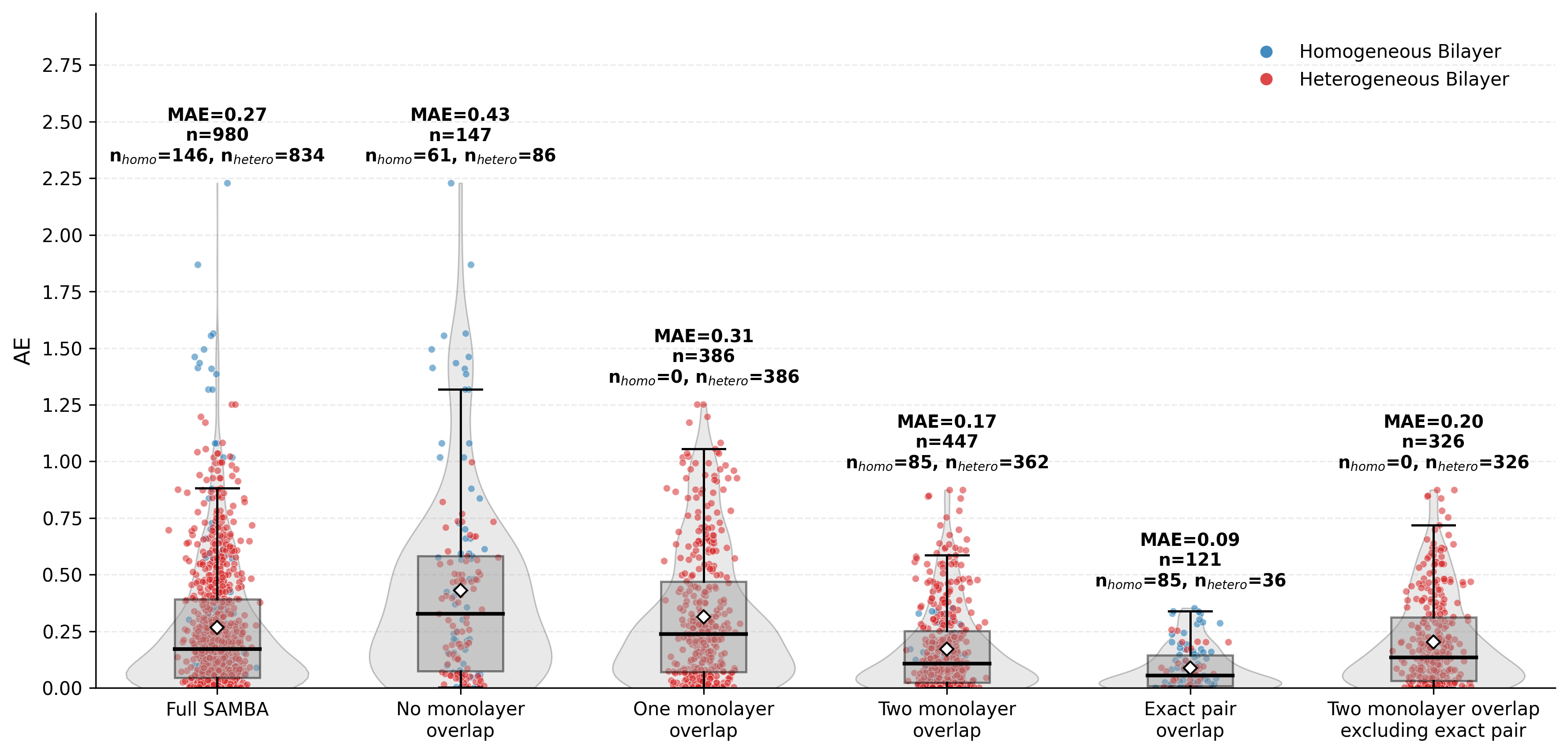}
    \caption{BiDB$^{*}$+HetDB-to-SAMBA Domain Generalization Performance Breakdown}
    \label{fig:box_plot_bidb_star_hetdb_samba_mono}
\end{figure}

\begin{figure}[t]
    \centering
    \includegraphics[width=1\linewidth]{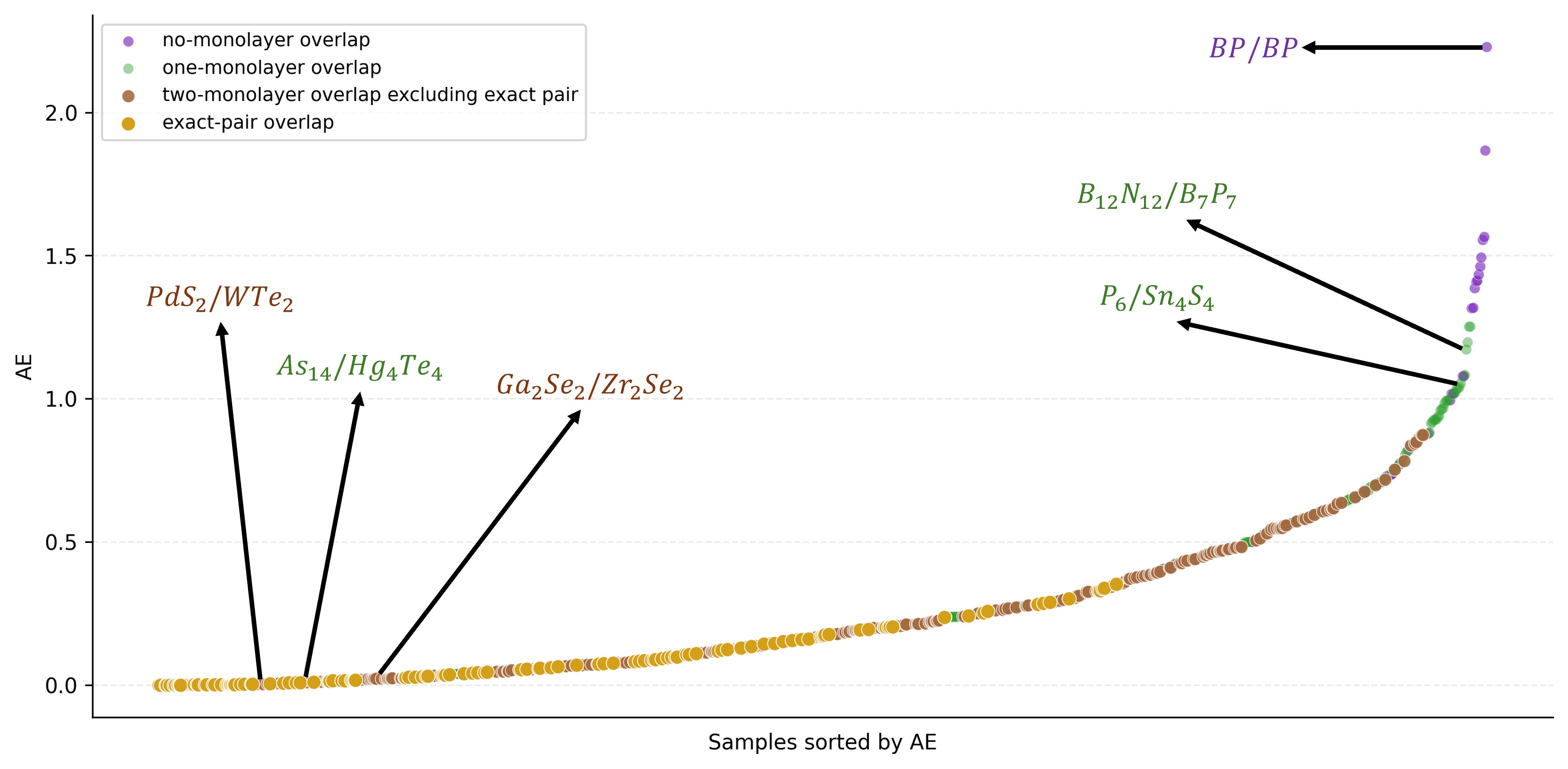}
    \caption{BiDB$^{*}$+HetDB-to-SAMBA Domain Generalization AE Distribution}
    \label{fig:scatter_plot_bidb_star_hetdb_samba_mono}
\end{figure}

\noindent\textbf{Prediction Performance.}
As shown in Table~\ref{tab:cross_domain_generalization_bidb_star_hetdb_samba}, BDIP-Net achieves better prediction performance than PotNet on the full SAMBA test set of 980 bilayers, with an MAE of \(0.27\), an MSE of \(0.16\), an RMSE of \(0.40\), and an \(R^2\) of \(0.40\), compared with \(0.34\), \(0.22\), \(0.47\), and \(0.15\), respectively, for PotNet. Figs.~\ref{fig:box_plot_bidb_star_hetdb_samba_mono} and~\ref{fig:scatter_plot_bidb_star_hetdb_samba_mono} show that the prediction performance of both models improves as the degree of monolayer overlap increases. On the 147 no-monolayer-overlap bilayers, BDIP-Net and PotNet achieve MAEs of \(0.43\) and \(0.48\), respectively. On the 386 one-monolayer-overlap bilayers, the corresponding MAEs are \(0.31\) and \(0.41\). On the 447 two-monolayer-overlap bilayers, BDIP-Net achieves an MAE of \(0.17\), compared with \(0.23\) for PotNet. The best results are obtained on the 121 exact-pair-overlap bilayers, where both models achieve an MAE of \(0.09\). After removing these exact-pair-overlap bilayers, BDIP-Net still achieves an MAE of \(0.20\), an RMSE of \(0.29\), and an \(R^2\) of \(0.54\) on the remaining 326 two-monolayer-overlap bilayers, compared with \(0.29\), \(0.40\), and \(0.12\), respectively, for PotNet. These results demonstrate that BDIP-Net provides stronger cross-dataset generalization to unseen bilayer combinations of known monolayers.

\noindent\textbf{Representative Prediction Cases.}
Table~\ref{tab:representative_overlap_bilayers} presents representative bilayers 
to illustrate how prediction difficulty varies with monolayer overlap and the relationship between bilayer and monolayer bandgaps. In the two-monolayer-overlap cases, BDIP-Net is more accurate than PotNet whether the bilayer bandgap is close to one monolayer bandgap or differs from both. For one-monolayer-overlap cases, BDIP-Net performs well when the bilayer bandgap is close to the overlapping monolayer bandgap, while both models show larger errors when it is closer to the non-overlapping monolayer bandgap or differs from both. In the no-monolayer-overlap case, both models have large prediction errors.

\begin{table}[t]
\centering
\caption{Representative bilayers with different degrees of monolayer overlap in the BiDB$^{*}$+HetDB-to-SAMBA cross-domain generalization experiment.}
\label{tab:representative_overlap_bilayers}

\resizebox{\columnwidth}{!}{%
\begin{tabular}{lcccccccc}
\toprule
\textbf{Bilayer} &
\textbf{\makecell{Bottom\\Layer}} &
\textbf{\makecell{Top\\Layer}} &
\textbf{\makecell{Twist\\Angle}} &
\textbf{\makecell{Bottom\\Bandgap}} &
\textbf{\makecell{Top\\Bandgap}} &
\textbf{\makecell{Bilayer\\Bandgap}} &
\textbf{\makecell{BDIP-Net\\Prediction (AE)}} &
\textbf{\makecell{PotNet\\Prediction (AE)}} \\
\midrule

Ga$_2$Se/$_2$ZrSe$_2$
& \underline{Ga$_2$Se$_2$}
& \underline{ZrSe$_2$}
& $0.00^\circ$
& 2.02
& 0.20
& 0.21
& 0.24 (0.03)
& 0.55 (0.34) \\

PdS$_2$/WTe$_2$
& \underline{PdS$_2$}
& \underline{WTe$_2$}
& $60.00^\circ$
& 1.17
& 0.81
& 0.13
& 0.12 (0.01)
& 0.32 (0.19) \\

As$_{14}/$Hg$_4$Te$_4$
& As$_2$
& \underline{HgTe}
& $19.11^\circ$
& 1.39
& 0.05
& 0.08
& 0.07 (0.01)
& 0.45 (0.37) \\

B$_{12}$N$_{12}/$B$_7$P$_7$
& \underline{BN}
& BP
& $49.11^\circ$
& 4.68
& 0.91
& 0.91
& 2.08 (1.17)
& 2.59 (1.68) \\

P$_6$/Sn$_4$S$_4$
& P$_2$
& \underline{SnS}
& $30.00^\circ$
& 1.98
& 2.32
& 1.70
& 0.61 (1.09)
& 0.81 (0.89) \\

BP/BP
& BP
& BP
& $60.00^\circ$
& 0.91
& 0.91
& 0.40
& 2.63 (2.23)
& 2.64 (2.24) \\

\bottomrule
\end{tabular}%
}

\vspace{1pt}
\parbox{\columnwidth}{\footnotesize
\textit{Note:} Underlined bottom- and top-layer monolayers overlap with monolayers in the training dataset.
}
\end{table}
\noindent\textbf{HetDB-to-SAMBA.}
We further evaluate the more challenging HetDB-to-SAMBA setting, where only the 336 HetDB heterobilayers are used for training. Since the number of monolayers overlapping with SAMBA decreases from 39 to 13, the overall performance of both BDIP-Net and PotNet degrades substantially. Nevertheless, BDIP-Net still generalizes better than PotNet to unseen bilayer combinations of known monolayers, particularly on the two-monolayer-overlap subset after excluding exact-pair overlap. Detailed results are provided in Appendix~\ref{app:hetdb_to_samba}.

\section{Conclusion}
In this work, we developed a pipeline for stacked bilayer material property prediction, from bilayer structure construction and efficient MatterSim-D3-based structural optimization to ML-based property prediction. Compared with Bimat-ML, which directly predicts bilayer properties from monolayer structures and stacking information, our workflow includes an additional bilayer structure construction and structural optimization step. However, this step is performed efficiently using MatterSim-D3, making the overall bilayer construction and optimization process significantly faster than workflows based on full DFT-based structural optimization. Moreover, the reconstructed bilayer graph enables GNN-based models to explicitly learn intra-layer and inter-layer atomic interactions. We further proposed BDIP-Net, an interaction-aware GNN that separates intra-layer and inter-layer interactions and fuses their messages through the interaction-message attention module. Experiments on BiDB, HetDB, and SAMBA show that BDIP-Net consistently outperforms the baseline models under both structural optimization settings. In addition, when bilayer structures optimized by MatterSim-D3 are used as model inputs, the prediction performance is nearly identical to that obtained using the corresponding DFT-optimized structures. The cross-dataset generalization experiments further show that BDIP-Net provides stronger performance than PotNet on unseen bilayer combinations formed from monolayers observed during training, while prediction becomes more challenging when one or both target monolayers are absent from the training data. These results demonstrate that the proposed workflow substantially reduces the structural optimization cost while maintaining accurate property prediction and supporting cross-dataset generalization for stacked bilayer materials.

\section*{Code Availability}
The source code and implementation details of BDIP-Net are publicly available at
\url{https://github.com/AnVuong123/BDIPNet}.

\section*{Limitations and Ethical Considerations}
This work is evaluated on publicly available computational materials databases and therefore inherits any biases, inaccuracies, or limited material diversity present in these datasets, which may affect model generalization. Although the proposed framework substantially reduces the computational cost of bilayer structure generation and property prediction, its predictions should complement rather than replace high-fidelity first-principles calculations and experimental validation. The study does not involve human participants or personal data; consequently, issues of data privacy and informed consent are not applicable. 

\section*{Generative AI Usage}
During the preparation of this manuscript, the authors used ChatGPT solely to assist with language editing. The generative AI tool was not used for the conception of the research, experimental design, data analysis, interpretation of results, or generation of scientific content. All scientific content, analyses, and conclusions were developed and verified by the authors, who take full responsibility for the accuracy and integrity of the work.

\newpage
\section*{Acknowledgements}
This work was supported in part by the National Institute of General Medical Sciences of National Institutes of Health under award P20GM139768, and the Arkansas Integrative Metabolic Research Center at the University of Arkansas.
\bibliographystyle{ACM-Reference-Format}
\bibliography{sample-base}


\appendix

\section{Background on Stacked 2D Materials Representation}
\label{app:background}

A crystal structure is represented as a periodic repetition of a unit cell in three-dimensional (3D) space. Specifically, a crystal is described by $M = (A, P, L)$, where $A = [\mathbf{a}_1, \dots, \mathbf{a}_N]^T \in \mathbb{R}^{N \times d_a}$ is the atom feature matrix, $P = [\mathbf{p}_1, \dots, \mathbf{p}_N]^T \in \mathbb{R}^{N \times 3}$ is the matrix of Cartesian coordinates of atoms in the unit cell, and $L = [\mathbf{l}_1, \mathbf{l}_2, \mathbf{l}_3]^T \in \mathbb{R}^{3 \times 3}$ is the lattice matrix describing how the unit cell repeats in three directions through the lattice vectors $\mathbf{l}_1, \mathbf{l}_2, \mathbf{l}_3$. The lattice vectors define the lattice parameters $a = \|\mathbf{l}_1\|$, $b = \|\mathbf{l}_2\|$, $c = \|\mathbf{l}_3\|$, and angles $\gamma = \angle(\mathbf{l}_1, \mathbf{l}_2)$, $\alpha = \angle(\mathbf{l}_2, \mathbf{l}_3)$, $\beta = \angle(\mathbf{l}_1, \mathbf{l}_3)$. 
The infinite crystal structure can then be represented as
\[
\hat{P} = \left\{ \hat{\mathbf{p}}_i \,\middle|\, \hat{\mathbf{p}}_i = \mathbf{p}_i + k_1 \mathbf{l}_1 + k_2 \mathbf{l}_2 + k_3 \mathbf{l}_3,\; k_1, k_2, k_3 \in \mathbb{Z},\; 1 \le i \le N \right\},
\]
\[
\hat{A} = \left\{ \hat{\mathbf{a}}_i \,\middle|\, \hat{\mathbf{a}}_i = \mathbf{a}_i,\; 1 \le i \le N \right\}.
\]

Here, $\hat{P}$ contains all periodic images of atomic positions generated by lattice translations, while $\hat{A}$ assigns the corresponding atomic features to each periodic image.

A two-dimensional (2D) material in the Computational 2D Materials Database (C2DB) 
is represented within this framework as a crystal structure with periodicity restricted to the in-plane directions and is described by $M = (A, P, L)$. The lattice is defined by the vectors $\mathbf{l}_1, \mathbf{l}_2, \mathbf{l}_3$, where $\mathbf{l}_1$ and $\mathbf{l}_2$ span the in-plane Bravais lattice, while the out-of-plane vector satisfies $\alpha = \beta = 90^\circ$ and $c = t + v$, such that $\mathbf{l}_3 = (0, 0, c)$, where $t$ denotes the thickness of the material and $v$ is a vacuum region introduced to prevent spurious interactions between periodic images along the out-of-plane direction. The atomic positions are given by the Cartesian coordinate matrix $P = [\mathbf{p}_1, \dots, \mathbf{p}_N]^T$. The infinite 2D material structure can then be represented as
\[
\hat{P} = \left\{ \hat{\mathbf{p}}_i \,\middle|\, \hat{\mathbf{p}}_i = \mathbf{p}_i + k_1 \mathbf{l}_1 + k_2 \mathbf{l}_2,\; k_1, k_2 \in \mathbb{Z},\; 1 \le i \le N \right\},
\]
\[
\hat{A} = \left\{ \hat{\mathbf{a}}_i \,\middle|\, \hat{\mathbf{a}}_i = \mathbf{a}_i,\; 1 \le i \le N \right\}.
\]

Here, $\hat{P}$ contains all periodic images of atomic positions generated by in-plane lattice translations, while $\hat{A}$ assigns the corresponding atomic features to each periodic image.

A stacked two-dimensional material in BiDB is constructed by combining two identical monolayer structures to form a homogeneous bilayer. Let $M^{\mathrm{mono}} = (A^{\mathrm{mono}}, P^{\mathrm{mono}}, L^{\mathrm{mono}})$ denote the monolayer structure, and define the bottom and top layers as $M^{\mathrm{bot}}$ and $M^{\mathrm{top}}$, respectively. Prior to stacking, a set of transformations may be applied to $M^{\mathrm{top}}$, including in-plane rotation, translation, and an optional flip operation. These transformations are parameterized by a stacking configuration matrix $S$ and applied to the atomic positions of the top layer as $P^{\mathrm{top}} \leftarrow P^{\mathrm{top}} S$. The two layers are then stacked along the out-of-plane direction. The optimal interlayer distance $d_{*}$ is selected by minimizing the total energy of the stacked structure along the out-of-plane direction. The bilayer lattice is constructed by preserving the in-plane lattice vectors and redefining the out-of-plane lattice vector as $L^{\mathrm{bi}} = [\mathbf{l}_1^{\mathrm{mono}}, \mathbf{l}_2^{\mathrm{mono}}, \mathbf{l}_3^{\mathrm{bi}}]^T$, where $\mathbf{l}_3^{\mathrm{bi}} = (0,0,c^{\mathrm{bi}})$ and $c^{\mathrm{bi}} = 2t^{\mathrm{mono}} + d^{} + v$. After stacking with the optimized interlayer distance, the bilayer structure is relaxed to optimize the atomic positions in all three spatial directions ($x$, $y$, and $z$), allowing further reduction of the total energy and yielding a stable bilayer material. The resulting bilayer is represented as $M^{\mathrm{bi}} = (A^{\mathrm{bi}}, P^{\mathrm{bi}}, L^{\mathrm{bi}})$, where $A^{\mathrm{bi}}$ and $P^{\mathrm{bi}}$ are obtained by combining the atomic features and positions of $M^{\mathrm{bot}}$ and $M^{\mathrm{top}}$.

For HetDB and SAMBA, the bilayer structure is constructed using a commensurate-supercell procedure. Let $M^{\mathrm{top}} = (A^{\mathrm{top}}, P^{\mathrm{top}}, L^{\mathrm{top}})$ and $M^{\mathrm{bot}} = (A^{\mathrm{bot}}, P^{\mathrm{bot}}, L^{\mathrm{bot}})$ denote the two input monolayers, which may be identical for homobilayers or different for heterobilayers. Given a pair of commensurate supercell matrices $S^{\mathrm{top}}$ and $S^{\mathrm{bot}}$, selected according to the dataset-specific lattice matching criteria, the corresponding supercells $M^{\mathrm{top,sc}}$ and $M^{\mathrm{bot,sc}}$ are first generated. The in-plane lattice vectors of the two supercells are then aligned by averaging the in-plane components of $L^{\mathrm{top,sc}}$ and $L^{\mathrm{bot,sc}}$ to define a common bilayer lattice, with $\mathbf{l}_1^{\mathrm{bi}} = \frac{1}{2}(\mathbf{l}_1^{\mathrm{top,sc}} + \mathbf{l}_1^{\mathrm{bot,sc}})$ and $\mathbf{l}_2^{\mathrm{bi}} = \frac{1}{2}(\mathbf{l}_2^{\mathrm{top,sc}} + \mathbf{l}_2^{\mathrm{bot,sc}})$. The bilayer lattice is defined as $L^{\mathrm{bi}} = [\mathbf{l}_1^{\mathrm{bi}}, \mathbf{l}_2^{\mathrm{bi}}, \mathbf{l}_3^{\mathrm{bi}}]^T$, where $\mathbf{l}_3^{\mathrm{bi}} = (0,0, v + d + t^{\mathrm{mono,top}} + t^{\mathrm{mono,bot}})$. The two supercells are then stacked along the out-of-plane direction. The optimal interlayer distance $d_{*}$ is first selected by minimizing the total energy of the stacked structure along the out-of-plane direction. Depending on the dataset-specific construction workflow, an in-plane $xy$-scan may also be performed after determining $d_{*}$. The in-plane stacking configuration is then selected as the lateral shift that gives the lowest total energy of the bilayer structure. After that the bilayer is structurally relaxed to optimize the atomic positions in all three spatial directions ($x$, $y$, and $z$), yielding a stable bilayer material. The final bilayer is represented as $M^{\mathrm{bi}} = (A^{\mathrm{bi}}, P^{\mathrm{bi}}, L^{\mathrm{bi}})$, where $A^{\mathrm{bi}}$ and $P^{\mathrm{bi}}$ are obtained by combining the atomic features and positions of $M^{\mathrm{bot,sc}}$ and $M^{\mathrm{top,sc}}$.

\section{Initial Bilayer Structure Construction}
\label{app:bilayer_construction}

This appendix provides the detailed process used to construct the initial stacked bilayer structures for BiDB, HetDB, and SAMBA before structural optimization.

\subsection{BiDB}
\label{app:bidb_construction}

\begin{algorithm}[t]
\caption{Initial Bilayer Construction in BiDB $(\mathrm{ConstructBIDB})$}
\label{alg:bilayer_construction}
\begin{algorithmic}[1]
\Input Monolayer crystal $M^{\mathrm{mono}} = (A^{\mathrm{mono}}, P^{\mathrm{mono}}, L^{\mathrm{mono}})$, bottom and top stacking configuration matrices $S^{\mathrm{bot}}$ and $S^{\mathrm{top}}$, initial interlayer distance $d$, in-plane translations $(\delta_x,\delta_y)$, monolayer thickness $t^{\mathrm{mono}}$, vacuum $v$
\Output Initial bilayer structure $M_{\mathrm{init}}^{\mathrm{bi}}$

\State Define the bottom and top stacking configuration matrices:
\[
S^{\mathrm{bot}}
=
\begin{bmatrix}
c_1^{\mathrm{bot}} & c_3^{\mathrm{bot}} & 0 \\
c_2^{\mathrm{bot}} & c_4^{\mathrm{bot}} & 0 \\
0 & 0 & c_5^{\mathrm{bot}}
\end{bmatrix},
\qquad
S^{\mathrm{top}}
=
\begin{bmatrix}
c_1^{\mathrm{top}} & c_3^{\mathrm{top}} & 0 \\
c_2^{\mathrm{top}} & c_4^{\mathrm{top}} & 0 \\
0 & 0 & c_5^{\mathrm{top}}
\end{bmatrix}
\]

\State Construct the bottom and top monolayer copies:
\[
M^{\mathrm{bot}} = M^{\mathrm{mono}}, \qquad
M^{\mathrm{top}} = M^{\mathrm{mono}}
\]

\State Compute the fractional coordinates of the bottom and top layers:
\[
F^{\mathrm{bot}} = P^{\mathrm{mono}}(L^{\mathrm{mono}})^{-1},
\qquad
F^{\mathrm{top}} = P^{\mathrm{mono}}(L^{\mathrm{mono}})^{-1}
\]

\State Apply the bottom and top stacking configuration matrices:
\[
F^{\mathrm{bot}} = (F^{\mathrm{bot}} S^{\mathrm{bot}})\bmod 1,
\qquad
F^{\mathrm{top}} = (F^{\mathrm{top}} S^{\mathrm{top}}) \bmod 1
\]

\State Define the bilayer thickness and lattice matrix:
\[
t^{\mathrm{bi}} = 2t^{\mathrm{mono}} + d, \quad
\mathbf{l}_3^{\mathrm{bi}} = (0,0,\, v + t^{\mathrm{bi}}), \quad
L^{\mathrm{bi}} =
\begin{bmatrix}
\mathbf{l}_1^{\mathrm{mono}} ,
\mathbf{l}_2^{\mathrm{mono}} ,
\mathbf{l}_3^{\mathrm{bi}}
\end{bmatrix} ^T
\in \mathbb{R}^{3 \times 3}
\]

\State Calculate the fractional out-of-plane layer translation:
\[
\delta z =
z_{\max}^{\mathrm{bot}}
+
\dfrac{d}{\|\mathbf{l}_3^{\mathrm{bi}}\|}
-
z_{\min}^{\mathrm{top}}
\]

\State Define the layer translation vector $\boldsymbol{\tau}$:
\[
\boldsymbol{\tau} =
\begin{bmatrix}
\delta_x & \delta_y & \delta z
\end{bmatrix}
\]

\State Apply the layer translation to the top layer:
\[
F^{\mathrm{top}} =
\left(
F^{\mathrm{top}} + \mathbf{1}\boldsymbol{\tau}
\right)\bmod 1,
\quad
\mathbf{1} \in \mathbb{R}^{N \times 1}
\]

\State Construct the bilayer fractional coordinates:
\[
F^{\mathrm{bi}} = F^{\mathrm{bot}} \cup F^{\mathrm{top}}
\]

\State Convert the bilayer fractional coordinates back to Cartesian coordinates:
\[
P^{\mathrm{bi}} = F^{\mathrm{bi}} L^{\mathrm{bi}}
\]

\State Construct the bilayer atomic features:
\[
A^{\mathrm{bi}} = A^{\mathrm{mono}} \cup A^{\mathrm{mono}}
\]

\State \Return $M_{\mathrm{init}}^{\mathrm{bi}} = (A^{\mathrm{bi}}, P^{\mathrm{bi}}, L^{\mathrm{bi}})$
\end{algorithmic}
\end{algorithm}

The initial bilayer construction workflow in the BiDB dataset follows Algorithm~\ref{alg:bilayer_construction}. Given a monolayer crystal $M^{\mathrm{mono}}$, bottom and top stacking configuration matrices $S^{\mathrm{bot}}$ and $S^{\mathrm{top}}$, an initial interlayer distance $d$, an in-plane translation $(\delta_x,\delta_y)$, and the monolayer thickness $t^{\mathrm{mono}}$, the goal is to construct the corresponding initial bilayer structure $M_{\mathrm{init}}^{\mathrm{bi}}$, where$ M^{\mathrm{mono}} = (A^{\mathrm{mono}},P^{\mathrm{mono}},L^{\mathrm{mono}})$, $M_{\mathrm{init}}^{\mathrm{bi}} =(A^{\mathrm{bi}},P^{\mathrm{bi}},L^{\mathrm{bi}}).$

The stacking configuration matrices for the bottom and top layers are written as
\begin{equation}
\label{eq:bidb_stack_matrices}
S^{\mathrm{bot}} =
\begin{bmatrix}
c_1^{\mathrm{bot}} & c_3^{\mathrm{bot}} & 0 \\
c_2^{\mathrm{bot}} & c_4^{\mathrm{bot}} & 0 \\
0 & 0 & c_5^{\mathrm{bot}}
\end{bmatrix},
\qquad
S^{\mathrm{top}} =
\begin{bmatrix}
c_1^{\mathrm{top}} & c_3^{\mathrm{top}} & 0 \\
c_2^{\mathrm{top}} & c_4^{\mathrm{top}} & 0 \\
0 & 0 & c_5^{\mathrm{top}}
\end{bmatrix},
\end{equation}
where $c_1,c_2,c_3,c_4$ define the in-plane transformation and $c_5$ defines the out-of-plane flip operation. In the current version of BiDB, the bottom layer is kept unchanged, while the stacking information is applied to the top layer. Therefore, the bottom-layer stacking configuration matrix is set to
\[
S^{\mathrm{bot}} = I_3 =
\begin{bmatrix}
1 & 0 & 0 \\
0 & 1 & 0 \\
0 & 0 & 1
\end{bmatrix}.
\]

The user-provided stacking information consists of the top-layer stacking configuration matrix $S^{\mathrm{top}}$ and the in-plane translations $(\delta_x,\delta_y)$. From the input monolayer, two monolayer copies are first constructed as $M^{\mathrm{bot}} = M^{\mathrm{mono}}$ and $M^{\mathrm{top}} = M^{\mathrm{mono}}$. The atomic positions of both copies are then converted into fractional coordinates with respect to the monolayer lattice, where $F^{\mathrm{bot}} = P^{\mathrm{mono}}(L^{\mathrm{mono}})^{-1}$ and $F^{\mathrm{top}} = P^{\mathrm{mono}}(L^{\mathrm{mono}})^{-1}$.
The bottom and top stacking configuration matrices are applied separately, followed by modulo 1:
\begin{equation}
\label{eq:bidb_apply_stack_matrices}
F^{\mathrm{bot}} =
\left(F^{\mathrm{bot}}S^{\mathrm{bot}}\right)\bmod 1,
\qquad
F^{\mathrm{top}} =
\left(F^{\mathrm{top}}S^{\mathrm{top}}\right)\bmod 1.
\end{equation}
Thus, the bottom layer remains unchanged, while the top layer is transformed according to the user-provided $S^{\mathrm{top}}$. The bilayer thickness is first defined as $t^{\mathrm{bi}} = 2t^{\mathrm{mono}} + d$. The bilayer out-of-plane lattice vector is then given by $\mathbf{l}_3^{\mathrm{bi}} = (0,0,v+t^{\mathrm{bi}})$, and the bilayer lattice matrix is defined as $L^{\mathrm{bi}} = \begin{bmatrix} \mathbf{l}_1^{\mathrm{mono}}, \mathbf{l}_2^{\mathrm{mono}}, \mathbf{l}_3^{\mathrm{bi}} \end{bmatrix}^{T}$.
The fractional out-of-plane layer translation $\delta z$ is computed to impose the initial interlayer distance $d$. First, the top surface of the bottom layer and the bottom surface of the top layer are computed as
\begin{equation}
\label{eq:bidb_zmax_zmin}
z_{\max}^{\mathrm{bot}}
=
\max \left( f_1^{3,\mathrm{bot}}, \dots, f_N^{3,\mathrm{bot}} \right),
\qquad
z_{\min}^{\mathrm{top}}
=
\min \left( f_1^{3,\mathrm{top}}, \dots, f_N^{3,\mathrm{top}} \right).
\end{equation}
Then, the fractional out-of-plane translation is given by
\begin{equation}
\label{eq:bidb_delta_z}
\delta z =
z_{\max}^{\mathrm{bot}}
+
\dfrac{d}{\|\mathbf{l}_3^{\mathrm{bi}}\|}
-
z_{\min}^{\mathrm{top}},
\end{equation}
where $f_i^{\mathrm{bot}} = (f_i^{1,\mathrm{bot}}, f_i^{2,\mathrm{bot}}, f_i^{3,\mathrm{bot}})$ and $f_i^{\mathrm{top}} = (f_i^{1,\mathrm{top}}, f_i^{2,\mathrm{top}}, f_i^{3,\mathrm{top}})$ denote the fractional coordinates of atom $i$ in the bottom and top layers, respectively. The layer translation vector is then defined using the user-provided in-plane translations and the computed out-of-plane translation:
\begin{equation}
\label{eq:bidb_layer_translation_vector}
\boldsymbol{\tau}
=
\begin{bmatrix}
\delta_x & \delta_y & \delta z
\end{bmatrix}.
\end{equation}
The layer translation is applied only to the top layer:
\begin{equation}
\label{eq:bidb_apply_layer_translation}
F^{\mathrm{top}}
=
\left(
F^{\mathrm{top}}
+
\mathbf{1}\boldsymbol{\tau}
\right)
\bmod 1,
\qquad
\mathbf{1}\in\mathbb{R}^{N\times 1}.
\end{equation}

The bilayer fractional coordinates are constructed by combining the bottom layer and the translated top layer: $F^{\mathrm{bi}} = F^{\mathrm{bot}} \cup F^{\mathrm{top}}$. These fractional coordinates are converted back to Cartesian coordinates using the bilayer lattice: $P^{\mathrm{bi}} = F^{\mathrm{bi}}L^{\mathrm{bi}}$. Finally, the bilayer atomic features are constructed as $A^{\mathrm{bi}} = A^{\mathrm{mono}} \cup A^{\mathrm{mono}}$, and the initial bilayer structure is given by $M_{\mathrm{init}}^{\mathrm{bi}} = (A^{\mathrm{bi}},P^{\mathrm{bi}},L^{\mathrm{bi}})$.

\subsection{HetDB}
\label{app:hetdb_construction}

The HetDB initial bilayer construction consists of two main steps. First, an optimal pair of compatible two-dimensional supercell transformation matrices, $C^{\mathrm{top}}$ and $C^{\mathrm{bot}}$, is identified for the two input monolayers using Algorithm~\ref{alg:supercell_pair_det}. Second, the selected two-dimensional matrix pair is used as input to Algorithm~\ref{alg:hetdb_bilayer_construction}, where it is extended to the corresponding three-dimensional supercell transformation matrices, $S^{\mathrm{top}}$ and $S^{\mathrm{bot}}$, by adding the out-of-plane flip operation. The resulting three-dimensional transformation matrix pair is then used to construct the initial heterogeneous bilayer structure $M_{\mathrm{init}}^{\mathrm{bi}}$.

\begin{algorithm}[t]
\caption{Supercell Pair Search in HetDB Dataset $(\mathrm{SupercellPairSearch})$}
\label{alg:supercell_pair_det}
\begin{algorithmic}[1]

\Input 
Monolayers $M^{\mathrm{top}}, M^{\mathrm{bot}}$; 
search bound $c_{\max}$

\Output Optimal pair $(C^{\mathrm{top}}_{\mathrm{opt}}, C^{\mathrm{bot}}_{\mathrm{opt}})$

\State Generate $\mathcal{C}^{(n)}$ using Eq.~\eqref{eq:hetdb_2d_supercell_transformation_matrix}, where the entries $c_i^{(n)}$ of each $C^{(n)}$ are constrained by $c_{\max}$, $n \in \{\mathrm{top}, \mathrm{bot}\}$

\State Sort $\mathcal{C}^{(n)}$ by increasing $|\det(C^{(n)})|$, $n \in \{\mathrm{top}, \mathrm{bot}\}$

\State Initialize $(C^{\mathrm{top}}_{\mathrm{opt}}, C^{\mathrm{bot}}_{\mathrm{opt}}) \leftarrow \emptyset$
\State Initialize $(N_{\mathrm{opt}}, \rho_{\mathrm{opt}}, \phi_{\mathrm{opt}}) \leftarrow (\infty, \infty, \infty)$

\For{$C^{\mathrm{bot}} \in \mathcal{C}^{\mathrm{bot}}$}
    \For{$C^{\mathrm{top}} \in \mathcal{C}^{\mathrm{top}}$}

        \State Compute total number of atoms $N$ using Eq.~\eqref{eq:hetdb_total_atoms}
        \If{$N > N_{\max}$}
            \State \textbf{break}
        \EndIf

        \State Construct supercell lattices $L^{(n),\mathrm{sc}}$ using Eq.~\eqref{eq:hetdb_lactice_definition}

        \State Compute twist angle $\phi$ using Eq.~\eqref{eq:hetdb_twisted_angle}

        \State Compute $\rho^{\mathrm{top}}, \rho^{\mathrm{bot}}$ and $\rho$ using Eq.~\eqref{eq:hetdb_internal_angle}

        \State Compute $\chi^{\mathrm{top}}, \chi^{\mathrm{bot}}$ using Eq.~\eqref{eq:hetdb_norm_ratio}

        \If{$\phi \notin [0^\circ, 90^\circ]$ \textbf{or}
            $\rho > \rho_{\max}$ \textbf{or}
            $\chi^{\mathrm{top}} \notin [15^\circ, 165^\circ]$ \textbf{or}
            $\chi^{\mathrm{bot}} \notin [15^\circ, 165^\circ]$}
            \State \textbf{continue}
        \EndIf
            
        \State Construct common lattice $L^{\mathrm{sc},\mathrm{common}}$ using Eq.~\eqref{eq:hetdb_common_lattice}

        \State Compute deformation tensors $s^{(n)}$ and strain tensors $\varepsilon^{(n)}$ using Eq.~\eqref{eq:hetdb_tensor}--\eqref{eq:hetdb_strain}

        \State Compute maximum strain $\varepsilon_{\text{max}}$ using Eq.~\eqref{eq:hetdb_max_strain}

        \If{$\varepsilon_{\text{max}} > \varepsilon_{\text{threshold}}$}
            \State \textbf{continue}
        \EndIf

        \If{$(N, \rho, \phi) < (N_{\mathrm{opt}}, \rho_{\mathrm{opt}}, \phi_{\mathrm{opt}})$}
            \State $(C^{\mathrm{top}}_{\mathrm{opt}}, C^{\mathrm{bot}}_{\mathrm{opt}}) \leftarrow (C^{\mathrm{top}}, C^{\mathrm{bot}})$
            \State $(N_{\mathrm{opt}}, \rho_{\mathrm{opt}}, \phi_{\mathrm{opt}}) \leftarrow (N, \rho, \phi)$
        \EndIf

    \EndFor
\EndFor

\State \Return $(C^{\mathrm{top}}_{\mathrm{opt}}, C^{\mathrm{bot}}_{\mathrm{opt}})$

\end{algorithmic}
\end{algorithm}

The supercell pair search in HetDB dataset follows Algorithm~\ref{alg:supercell_pair_det}. The procedure begins by extracting the primitive in-plane lattice matrices $L^{(n)} \in \mathbb{R}^{2\times2}$ for $n \in \{\mathrm{top}, \mathrm{bot}\}$ from the input monolayers, where $L^{(n)} = [\mathbf{l}_1^{(n)}, \mathbf{l}_2^{(n)}]^T$ contains only the in-plane lattice vectors, while the out-of-plane lattice vector $\mathbf{l}_3^{(n)}$ is retained but not used in the supercell pair search procedure. For each layer $n \in \{\mathrm{top}, \mathrm{bot}\}$, all possible integer two-dimensional supercell transformation matrices are generated as
\begin{equation}
\label{eq:hetdb_2d_supercell_transformation_matrix}
C^{(n)} =
\begin{pmatrix}
c_1^{(n)} & c_2^{(n)} \\
c_3^{(n)} & c_4^{(n)}
\end{pmatrix},
\quad
c_i^{(n)} \in \mathbb{Z}, \quad |c_i^{(n)}| \le c_{\max}.
\end{equation}
where $c_i^{(n)}$ are the integer coefficients defining the in-plane supercell transformation and $c_{\max}$ bounds their magnitude. This defines the candidate sets $\mathcal{C}^{(n)}$, $n \in \{\mathrm{top}, \mathrm{bot}\}$. The determinant $|\det(C^{(n)})|$ gives the scaling factor for the number of atoms, such that the supercell contains $|\det(C^{(n)})|$ times the number of atoms $N^{\mathrm{(n)}}_{\mathrm{mono}}$ in the monolayer unit cell. The candidate sets $\mathcal{C}^{(n)}$ are then sorted in increasing order of $|\det(C^{(n)})|$ to prioritize smaller supercells. For any candidate pair $(C^{\mathrm{top}}, C^{\mathrm{bot}})$, the total number of atoms is computed as
\begin{equation}
\label{eq:hetdb_total_atoms}
N = |\det(C^{\mathrm{top}})|\, N^{\mathrm{top}}_{\mathrm{mono}} 
+ |\det(C^{\mathrm{bot}})|\, N^{\mathrm{bot}}_{\mathrm{mono}}.
\end{equation}
Due to this ordering, for a fixed $C^{\mathrm{bot}}$ in the outer loop, the total number of atoms $N$ increases monotonically as $C^{\mathrm{top}}$ in the inner loop progresses. Therefore, if $N > N_{\max}$, the current candidate pair is discarded and the inner loop over $C^{\mathrm{top}}$ is terminated early.

For the remaining candidates, the corresponding in-plane supercell lattices are constructed as
\begin{equation}
\label{eq:hetdb_lactice_definition}
L^{(n),\mathrm{sc}} = C^{(n)} L^{(n)}, \quad 
L^{(n),\mathrm{sc}} = [\mathbf{l}_1^{(n),\mathrm{sc}}, \mathbf{l}_2^{(n),\mathrm{sc}}]^T,\quad
n \in \{\mathrm{top}, \mathrm{bot}\}.
\end{equation}
where $L^{(n),\mathrm{sc}} \in \mathbb{R}^{2\times2}$ denotes the supercell lattice of layer $n$. The twist angle is computed from the first lattice vectors of the two layers:
\begin{equation}
\label{eq:hetdb_twisted_angle}
\phi =
\tan^{-1}
\left(
\frac{
\mathbf{l}^{\mathrm{top},\mathrm{sc}}_1 \times \mathbf{l}^{\mathrm{bot},\mathrm{sc}}_1
}{
\mathbf{l}^{\mathrm{top},\mathrm{sc}}_1 \cdot \mathbf{l}^{\mathrm{bot},\mathrm{sc}}_1
}
\right).
\end{equation}

The internal angle $\chi^{(n)}$ is computed as
\begin{equation}
\label{eq:hetdb_internal_angle}
\chi^{(n)} =
\cos^{-1}
\left(
\frac{
\mathbf{l}^{(n),\mathrm{sc}}_1 \cdot \mathbf{l}^{(n),\mathrm{sc}}_2
}{
\|\mathbf{l}^{(n),\mathrm{sc}}_1\| \, \|\mathbf{l}^{(n),\mathrm{sc}}_2\|
}
\right).
\end{equation}

The lattice vector norm ratio $\rho^{(n)}$ is computed as
\begin{equation}
\label{eq:hetdb_norm_ratio}
\rho^{(n)} =
\max\left(
\frac{\|\mathbf{l}^{(n),\mathrm{sc}}_1\|}{\|\mathbf{l}^{(n),\mathrm{sc}}_2\|},
\frac{\|\mathbf{l}^{(n),\mathrm{sc}}_2\|}{\|\mathbf{l}^{(n),\mathrm{sc}}_1\|}
\right), \quad
\rho = \max(\rho^{\mathrm{top}}, \rho^{\mathrm{bot}}).
\end{equation}

Candidate pairs violating the geometric constraints $\phi \in [0^\circ, 90^\circ]$, $\chi^{\mathrm{top}}, \chi^{\mathrm{bot}} \in [15^\circ, 165^\circ]$, and $\rho \le \rho_{\max}$ are discarded. For valid candidates, a common in-plane lattice is constructed as
\begin{equation}
\label{eq:hetdb_common_lattice}
L^{\mathrm{sc},\mathrm{common}} = \frac{1}{2}\left(L^{\mathrm{top},\mathrm{sc}} + L^{\mathrm{bot},\mathrm{sc}}\right),
\end{equation}
which defines the shared in-plane lattice of the bilayer prior to introducing the out-of-plane lattice vector.

The deformation tensor of each layer is then computed by mapping its unstrained supercell lattice onto the common lattice:
\begin{equation}
\label{eq:hetdb_tensor}
\boldsymbol{s}^{(n)}
=
\left(L^{(n),\mathrm{sc}}\right)^{-1}
L^{\mathrm{sc},\mathrm{common}}
-
I,
\quad
n \in \{\mathrm{top}, \mathrm{bot}\},
\end{equation}
where \(I\) is the \(2\times2\) identity matrix. The corresponding finite Lagrangian strain tensor is defined as
\begin{equation}
\label{eq:hetdb_strain}
\boldsymbol{\varepsilon}^{(n)}
=
\frac{1}{2}
\left[
\boldsymbol{s}^{(n)}
+
\left(\boldsymbol{s}^{(n)}\right)^T
+
\boldsymbol{s}^{(n)}
\left(\boldsymbol{s}^{(n)}\right)^T
\right].
\end{equation}
The maximum strain of the candidate pair is obtained from the eigenvalues of the strain tensors:
\begin{equation}
\label{eq:hetdb_max_strain}
\varepsilon_{\max}
=
\max_{n \in \{\mathrm{top},\mathrm{bot}\}}
\max_i
\left|
\lambda_i\left(\boldsymbol{\varepsilon}^{(n)}\right)
\right|.
\end{equation}
If $\varepsilon_{\max} > \varepsilon_{\mathrm{threshold}}$, the candidate pair is discarded. Otherwise, it is compared with the current best solution using the lexicographic order of $(N,\rho,\phi)$, where smaller values are preferred. Finally, the best candidate pair is selected as $(C^{\mathrm{top}}_{\mathrm{opt}},C^{\mathrm{bot}}_{\mathrm{opt}})$.

\begin{algorithm}[t]
\caption{Initial Bilayer Construction in HetDB $(\mathrm{ConstructHETDB})$}
\label{alg:hetdb_bilayer_construction}
\begin{algorithmic}[1]

\Input 
Monolayers $M^{\mathrm{top}}, M^{\mathrm{bot}}$; 
two-dimensional supercell transformation matrices $C^{\mathrm{top}}, C^{\mathrm{bot}}$; 
flip operations $c_5^{\mathrm{top}}, c_5^{\mathrm{bot}}$; monolayer thicknesses $t^{\mathrm{mono,top}}, t^{\mathrm{mono,bot}}$; 
vacuum region $v$, interlayer distance $d$

\Output  Initial bilayer structure $M_{\mathrm{init}}^{\mathrm{bi}}$

\State Extend the two-dimensional supercell transformation matrices to three-dimensional matrices:
\[
C^{(n)} =
\begin{bmatrix}
c_1^{(n)} & c_2^{(n)} \\
c_3^{(n)} & c_4^{(n)}
\end{bmatrix},
\quad
S^{(n)} =
\begin{bmatrix}
c_1^{(n)} & c_2^{(n)} & 0 \\
c_3^{(n)} & c_4^{(n)} & 0 \\
0 & 0 & c_5^{(n)}
\end{bmatrix},
\quad
n \in \{\mathrm{top}, \mathrm{bot}\}.
\]

\State Construct supercells:
\[
M^{(n),\mathrm{sc}} = (A^{(n),\mathrm{sc}}, P^{(n),\mathrm{sc}}, L^{(n),\mathrm{sc}})
= \mathrm{MakeSupercell}(M^{(n)}, S^{(n)}).
\]

\State Compute common in-plane lattice $L^{\mathrm{sc},\mathrm{common}}$ using Eq.~\eqref{eq:hetdb_common_lattice}

\State Define bilayer lattice vectors:
\[
\mathbf{l}_1^{\mathrm{bi}} = \mathbf{l}_1^{\mathrm{sc},\mathrm{common}}, \quad
\mathbf{l}_2^{\mathrm{bi}} = \mathbf{l}_2^{\mathrm{sc},\mathrm{common}}, \quad
\mathbf{l}_3^{\mathrm{bi}} = (0, 0, v + d + t^{\mathrm{mono,top}} + t^{\mathrm{mono,bot}})
\]

\State Construct bilayer lattice:
\[
L^{\mathrm{bi}} =
\begin{bmatrix}
\mathbf{l}_1^{\mathrm{bi}},
\mathbf{l}_2^{\mathrm{bi}},
\mathbf{l}_3^{\mathrm{bi}}
\end{bmatrix}^T
\]

\State Convert to fractional coordinates:
\[
F^{(n),\mathrm{sc}} = P^{(n),\mathrm{sc}} (L^{\mathrm{bi}})^{-1},
\quad n \in \{\mathrm{top}, \mathrm{bot}\}
\]

\State Compute current interlayer distance:
\[
d_{\mathrm{current}}
=
\left(
z_{\mathrm{mean}}^{\mathrm{top}}
-
z_{\mathrm{mean}}^{\mathrm{bot}}
\right)
\lVert \mathbf{l}_3^{\mathrm{bi}} \rVert
\]

\State Compute fractional out-of-plane translation:
\[
\delta z = \frac{d - d_{\mathrm{current}}}{\lVert \mathbf{l}_3^{\mathrm{bi}} \rVert}
\]

\State Define the layer translation vector without in-plane translations:
\[
\boldsymbol{\tau}
=
\begin{bmatrix}
0 & 0 & \delta z
\end{bmatrix}
\]

\State Apply the layer translation to the top layer:
\[
F^{\mathrm{top},\mathrm{sc}} \leftarrow 
\left(
F^{\mathrm{top},\mathrm{sc}} + 
\mathbf{1}\boldsymbol{\tau}
\right) \bmod 1,
\qquad
\mathbf{1}\in\mathbb{R}^{N^{\mathrm{top},\mathrm{sc}}\times 1}
\]

\State Convert back to Cartesian coordinates:
\[
P^{(n),\mathrm{sc}} = F^{(n),\mathrm{sc}} L^{\mathrm{bi}}, \quad n \in \{\mathrm{top}, \mathrm{bot}\}
\]

\State Combine layers:
\[
P^{\mathrm{bi}} = P^{\mathrm{top},\mathrm{sc}} \cup P^{\mathrm{bot},\mathrm{sc}}, \quad
A^{\mathrm{bi}} = A^{\mathrm{top},\mathrm{sc}} \cup A^{\mathrm{bot},\mathrm{sc}}
\]

\State \Return $M^{\mathrm{bi}} = (A^{\mathrm{bi}}, P^{\mathrm{bi}}, L^{\mathrm{bi}})$

\end{algorithmic}
\end{algorithm}
The process of constructing an initial bilayer material in the HetDB dataset is summarized in Algorithm~\ref{alg:hetdb_bilayer_construction}. Given the top and bottom monolayers $M^{\mathrm{top}}$ and $M^{\mathrm{bot}}$, the two-dimensional supercell transformation matrices $C^{\mathrm{top}}$ and $C^{\mathrm{bot}}$, the flip operations $c_5^{\mathrm{top}}$ and $c_5^{\mathrm{bot}}$, the monolayer thicknesses, the vacuum spacing $v$, and the target interlayer distance $d$, the goal is to construct the heterogeneous bilayer material $M_{\mathrm{init}}^{\mathrm{bi}}$. The in-plane translations are fixed as $(\delta x,\delta y)=(0,0)$. We adopt the same definition of the flip operation as in BiDB, where the in-plane directions are preserved and the out-of-plane direction is either kept or reversed. In HetDB, two configurations are considered: keeping both layers unflipped or flipping both layers simultaneously.

The two-dimensional supercell transformation matrices are first extended to three-dimensional supercell transformation matrices by including the flip operation along the out-of-plane direction:
\begin{equation}
S^{(n)} =
\begin{bmatrix}
c_1^{(n)} & c_2^{(n)} & 0 \\
c_3^{(n)} & c_4^{(n)} & 0 \\
0 & 0 & c_5^{(n)}
\end{bmatrix}, \quad n \in \{\mathrm{top}, \mathrm{bot}\},
\end{equation}
where \(c_1^{(n)},c_2^{(n)},c_3^{(n)},c_4^{(n)}\) are from the two-dimensional supercell transformation matrix \(C^{(n)}\) and \(c_5^{(n)}=\pm 1\) controls the out-of-plane flip. The function $\mathrm{MakeSupercell}$ is then applied to each monolayer structure $M^{(n)}=(A^{(n)},P^{(n)},L^{(n)})$, where \(A^{(n)}\) is the atomic features, \(P^{(n)}\) contains the Cartesian atomic coordinates, and \(L^{(n)}\) is the lattice matrix. The corresponding supercell is constructed as
\begin{equation}
\label{eq:hetdb_make_supercell}
M^{(n),\mathrm{sc}}
=
(A^{(n),\mathrm{sc}},P^{(n),\mathrm{sc}},L^{(n),\mathrm{sc}})
=
\mathrm{MakeSupercell}(M^{(n)},S^{(n)}).
\end{equation}

Inside this function, the fractional coordinates are first computed from the Cartesian coordinates as
\begin{equation}
F^{(n)}=P^{(n)}\left(L^{(n)}\right)^{-1},
\end{equation}
and the supercell lattice is obtained as
\begin{equation}
L^{(n),\mathrm{sc}}=S^{(n)}L^{(n)}.
\end{equation}
The determinant of \(S^{(n)}\) gives the number of primitive cells contained in the supercell, so the number of atoms becomes
\begin{equation}
N^{(n),\mathrm{sc}}
=
\left|\det S^{(n)}\right|N^{(n)}
=
\left|
c_1^{(n)}c_4^{(n)}-c_2^{(n)}c_3^{(n)}
\right|N^{(n)},
\end{equation}
since \(c_5^{(n)}=\pm 1\) only changes the out-of-plane orientation. The supercell fractional coordinates are generated by translating each \(\mathbf{f}_i^{(n)}\in F^{(n)}\) with integer vectors \(\mathbf{t}=(t_1,t_2,t_3)\), where \(t_1,t_2,t_3\in\mathbb{Z}\). For a two-dimensional material, translations are applied only in the in-plane directions, so \(t_3=0\) and \(\mathbf{t}=(t_1,t_2,0)\). Only translated images inside the supercell are retained until \(N^{(n),\mathrm{sc}}\) atoms are generated:
\begin{equation}
F^{(n),\mathrm{sc}}
=
\left\{
\mathbf{f}_{i,\mathbf{t}}^{(n),\mathrm{sc}}
=
\left(\mathbf{f}_i^{(n)}+\mathbf{t}\right)
\left(S^{(n)}\right)^{-1}
\,\middle|\,
\mathbf{f}_{i,\mathbf{t}}^{(n),\mathrm{sc}}
\in [0,1)^3
\right\}.
\end{equation}

The retained fractional coordinates are converted back to Cartesian coordinates as
\begin{equation}
P^{(n),\mathrm{sc}}=F^{(n),\mathrm{sc}}L^{(n),\mathrm{sc}}.
\end{equation}
The corresponding atomic features are copied from $A^{(n)}$, forming $A^{(n),\mathrm{sc}}$. Next, the common in-plane lattice $L^{\mathrm{sc},\mathrm{common}}$ is computed from the top and bottom supercell lattices using Eq.~\eqref{eq:hetdb_common_lattice}. The bilayer out-of-plane length is first computed as $t^{\mathrm{bi}} = v+d+t^{\mathrm{mono,top}}+t^{\mathrm{mono,bot}}$, which includes the top-layer thickness, bottom-layer thickness, interlayer distance, and vacuum spacing. The bilayer lattice is then constructed by using the common lattice for the in-plane directions, with $\mathbf{l}_1^{\mathrm{bi}} = \mathbf{l}_1^{\mathrm{sc},\mathrm{common}}$, $\mathbf{l}_2^{\mathrm{bi}} = \mathbf{l}_2^{\mathrm{sc},\mathrm{common}}$, and $\mathbf{l}_3^{\mathrm{bi}} = (0,0,t^{\mathrm{bi}})$. The bilayer lattice matrix is then defined as $L^{\mathrm{bi}} = \begin{bmatrix} \mathbf{l}_1^{\mathrm{bi}}, \mathbf{l}_2^{\mathrm{bi}}, \mathbf{l}_3^{\mathrm{bi}} \end{bmatrix}^{T}$. The atomic positions of both supercells are converted into fractional coordinates with respect to the bilayer lattice:
\begin{equation}
F^{(n),\mathrm{sc}} = P^{(n),\mathrm{sc}}(L^{\mathrm{bi}})^{-1},
\quad n \in \{\mathrm{top}, \mathrm{bot}\}.
\end{equation}
The current interlayer distance is computed from the mean fractional $z$-coordinates of the top and bottom layers:
\begin{equation}
d_{\mathrm{current}}
=
\left(
z_{\mathrm{mean}}^{\mathrm{top}}
-
z_{\mathrm{mean}}^{\mathrm{bot}}
\right)
\lVert \mathbf{l}_3^{\mathrm{bi}} \rVert .
\end{equation}
To match the target interlayer distance $d$, the required fractional out-of-plane translation is
\begin{equation}
\delta z =
\frac{d-d_{\mathrm{current}}}
{\lVert \mathbf{l}_3^{\mathrm{bi}} \rVert}.
\end{equation}
The layer translation vector without in-plane translations is then defined as
\begin{equation}
\label{eq:translation_vector_no_inplane}
\boldsymbol{\tau}
=
\begin{bmatrix}
0 & 0 & \delta z
\end{bmatrix}.
\end{equation}
The layer translation is applied only to the top layer \(F^{\mathrm{top},\mathrm{sc}}\), following the same operation as in Eq.~\eqref{eq:bidb_apply_layer_translation}. Finally, the translated fractional coordinates are converted back to Cartesian coordinates as $P^{\mathrm{bot},\mathrm{sc}} = F^{\mathrm{bot},\mathrm{sc}}L^{\mathrm{bi}}$ and $P^{\mathrm{top},\mathrm{sc}} = F^{\mathrm{top},\mathrm{sc}}L^{\mathrm{bi}}$. The two layers are then combined as $P^{\mathrm{bi}} = P^{\mathrm{top},\mathrm{sc}} \cup P^{\mathrm{bot},\mathrm{sc}}$ and $A^{\mathrm{bi}} = A^{\mathrm{top},\mathrm{sc}} \cup A^{\mathrm{bot},\mathrm{sc}}$, yielding the initial heterogeneous bilayer structure $M_{\mathrm{init}}^{\mathrm{bi}} = (A^{\mathrm{bi}}, P^{\mathrm{bi}}, L^{\mathrm{bi}})$.

\subsection{SAMBA}
\label{app:samba_construction}

For a given pair of monolayers $M^{\mathrm{top}}$ and $M^{\mathrm{bot}}$, Algorithm~\ref{alg:samba_matching_pairs} summarizes the SAMBA initial bilayer construction. Candidate two-dimensional supercell transformation matrix pairs  are first generated and filtered according to geometric compatibility constraints, including lattice-area mismatch, internal-angle mismatch, and lattice-vector mismatch. Each retained two-dimensional matrix pair is then used as the input to Algorithm~\ref{alg:samba_bilayer_construction}, where it is extended to the corresponding three-dimensional supercell transformation matrix pair by adding the out-of-plane flip operation. The resulting three-dimensional transformation matrix pair is then used to construct a candidate initial bilayer structure. After all valid matrix pairs have been processed, duplicate bilayer structures are removed using RMSD-based filtering, yielding the final set of non-duplicated initial SAMBA bilayer structures $\mathcal{M}_{\mathrm{init}}^{\mathrm{bi}}$.

\begin{algorithm}[t]
\caption{Initial Bilayer Structures Generation in SAMBA}
\label{alg:samba_matching_pairs}
\begin{algorithmic}[1]

\Input 
Monolayers $M^{\mathrm{top}}, M^{\mathrm{bot}}$; 
search bound $c_{\max}$; 
thresholds $\epsilon_{\mathrm{area}}$, $\epsilon_\chi^{\%}$, $\epsilon_\chi^{\circ}$, $\epsilon_\ell$, initial interlayer distance $d$

\Output Set of initial bilayer structures $\mathcal{M}^{\mathrm{bi}}_{\mathrm{init}}$

\State Initialize $\mathcal{M}_{\mathrm{init}}^{\mathrm{bi}} \leftarrow \emptyset$
\State Generate candidate two-dimensional supercell transformation matrices $\mathcal{C}^{(n)}$ using Eq.~\eqref{eq:hetdb_2d_supercell_transformation_matrix}, with $|c_i^{(n)}| \le c_{\max}$

\For{$C^{\mathrm{bot}} \in \mathcal{C}^{\mathrm{bot}}$}
    \For{$C^{\mathrm{top}} \in \mathcal{C}^{\mathrm{top}}$}

        \State Construct supercell lattices $L^{(n),\mathrm{sc}}$ using Eq.~\eqref{eq:hetdb_lactice_definition}
        
        \State Compute area mismatch $\Delta_{\mathrm{area}}$ using Eq.~\eqref{eq:samba_area_mismatch}

        \State Compute internal angles $\chi^{\mathrm{top}}, \chi^{\mathrm{bot}}$ using Eq.~\eqref{eq:samba_internal_angle}
        
        \State Compute angle mismatches $\Delta_{\chi}^{\%}, \Delta_{\chi}^{\circ}$ using Eq.~\eqref{eq:samba_internal_angle_percentage}
        
        \State Compute lattice-vector mismatches $\Delta_{\mathbf{l},1}, \Delta_{\mathbf{l},2}$ using Eq.~\eqref{eq:samba_lattice_vector_mismatch}

        \If{
            $\Delta_{\mathrm{area}} > \epsilon_{\mathrm{area}}$
            \textbf{or}
            $\Delta_{\chi}^{\%} > \epsilon_\chi^{\%}$
            \textbf{or}
            $\Delta_{\chi}^{\circ} > \epsilon_\chi^{\circ}$
            \textbf{or}
            $\Delta_{\mathbf{l},1} > \epsilon_\ell$
            \textbf{or}
            $\Delta_{\mathbf{l},2} > \epsilon_\ell$
            }
                \State \textbf{continue}
            \EndIf

        \State Compute twist angle $\phi$ using Eq.~\eqref{eq:samba_twisted_angle}

        \State
        $
        M_{\mathrm{init}}^{\mathrm{bi}} \leftarrow 
        \mathrm{ConstructSAMBA}(
        M^{\mathrm{top}}, M^{\mathrm{bot}},
        C^{\mathrm{top}}, C^{\mathrm{bot}},\; d
        )
        $

        \State Add $M_{\mathrm{init}}^{\mathrm{bi}} $ to $\mathcal{M}_{\mathrm{init}}^{\mathrm{bi}}$

    \EndFor
\EndFor

\State Remove duplicate structures in $\mathcal{M}^{\mathrm{bi}}$ via RMSD-based filtering

\State \Return $\mathcal{M}_{\mathrm{init}}^{\mathrm{bi}}$

\end{algorithmic}
\end{algorithm}
Let $L^{\mathrm{top}} = [\mathbf{l}_1^{\mathrm{top}}, \mathbf{l}_2^{\mathrm{top}}]^T \in \mathbb{R}^{2\times2}$ and $L^{\mathrm{bot}} = [\mathbf{l}_1^{\mathrm{bot}}, \mathbf{l}_2^{\mathrm{bot}}]^T \in \mathbb{R}^{2\times2}$ denote the primitive in-plane lattice matrices of the top and bottom monolayers, respectively. For each monolayer pair, the initial bilayer candidate set is first initialized as $\mathcal{M}_{\mathrm{init}}^{\mathrm{bi}} \leftarrow \emptyset$. The candidate sets of two-dimensional supercell transformation matrices, $\mathcal{C}^{\mathrm{top}}$ and $\mathcal{C}^{\mathrm{bot}}$, are then generated using Eq.~\eqref{eq:hetdb_2d_supercell_transformation_matrix} with the search bound $c_{\max}$. For each matrix pair $C^{\mathrm{top}} \in \mathcal{C}^{\mathrm{top}}$ and $C^{\mathrm{bot}} \in \mathcal{C}^{\mathrm{bot}}$, the corresponding supercell lattice candidates $L^{\mathrm{top},\mathrm{sc}}$ and $L^{\mathrm{bot},\mathrm{sc}}$ are computed using Eq.~\eqref{eq:hetdb_lactice_definition}. For each candidate  pair, the area mismatch is defined as
\begin{equation}
\label{eq:samba_area_mismatch}
\Delta_{\mathrm{area}} =
\frac{
\left|
\|\mathbf{l}_1^{\mathrm{top},\mathrm{sc}} \times \mathbf{l}_2^{\mathrm{top},\mathrm{sc}}\|
-
\|\mathbf{l}_1^{\mathrm{bot},\mathrm{sc}} \times \mathbf{l}_2^{\mathrm{bot},\mathrm{sc}}\|
\right|
}{
\|\mathbf{l}_1^{\mathrm{top},\mathrm{sc}} \times \mathbf{l}_2^{\mathrm{top},\mathrm{sc}}\|
}.
\end{equation}

The internal angle of each supercell is defined as
\begin{equation}
\label{eq:samba_internal_angle}
\chi^{(n)} =
\cos^{-1}
\left(
\frac{
\mathbf{l}_1^{(n),\mathrm{sc}} \cdot \mathbf{l}_2^{(n),\mathrm{sc}}
}{
\|\mathbf{l}_1^{(n),\mathrm{sc}}\| \, \|\mathbf{l}_2^{(n),\mathrm{sc}}\|
}
\right).
\end{equation}

The relative and absolute internal angle mismatches are defined as
\begin{equation}
\label{eq:samba_internal_angle_percentage}
\Delta_{\chi}^{\%} =
\frac{|\chi^{\mathrm{top}} - \chi^{\mathrm{bot}}|}{\chi^{\mathrm{top}}},
\qquad
\Delta_{\chi}^{\circ} =
|\chi^{\mathrm{top}} - \chi^{\mathrm{bot}}|.
\end{equation}

The lattice-vector mismatch is defined as
\begin{equation}
\label{eq:samba_lattice_vector_mismatch}
\Delta_{\mathbf{l},i} =
\frac{
\left|
\|\mathbf{l}_i^{\mathrm{top},\mathrm{sc}}\|
-
\|\mathbf{l}_i^{\mathrm{bot},\mathrm{sc}}\|
\right|
}{
\|\mathbf{l}_i^{\mathrm{top},\mathrm{sc}}\|
}, \quad i \in \{1,2\}.
\end{equation}

Let $\epsilon_{\mathrm{area}}$, $\epsilon_\chi^{\%}$, $\epsilon_\chi^{\circ}$, and $\epsilon_\ell$ denote the corresponding thresholds. 
A candidate pair is discarded if
\begin{equation}
\Delta_{\mathrm{area}} > \epsilon_{\mathrm{area}}
\;\;\text{or}\;\;
\Delta_{\chi}^{\%} > \epsilon_\chi^{\%}
\;\;\text{or}\;\;
\Delta_{\chi}^{\circ} > \epsilon_\chi^{\circ}
\;\;\text{or}\;\;
\Delta_{\mathbf{l},1} > \epsilon_\ell
\;\;\text{or}\;\;
\Delta_{\mathbf{l},2} > \epsilon_\ell.
\end{equation}

Only pairs satisfying all constraints are retained. For these valid pairs, the twist angle is computed as
\begin{equation}
\label{eq:samba_twisted_angle}
\phi =
\tan^{-1}
\left(
\frac{
\mathbf{l}_1^{\mathrm{top},\mathrm{sc}} \times \mathbf{l}_1^{\mathrm{bot},\mathrm{sc}}
}{
\mathbf{l}_1^{\mathrm{top},\mathrm{sc}} \cdot \mathbf{l}_1^{\mathrm{bot},\mathrm{sc}}
}
\right).
\end{equation}

The retained two-dimensional matrix pairs $(C^{\mathrm{top}},C^{\mathrm{bot}})$ are then used as inputs to Algorithm~\ref{alg:samba_bilayer_construction}, where they are extended to the corresponding three-dimensional supercell transformation matrices $(S^{\mathrm{top}},S^{\mathrm{bot}})$. In the current SAMBA construction, the out-of-plane direction is kept unchanged for all materials. Therefore, the out-of-plane transformation is fixed as $c_5^{\mathrm{top}}=c_5^{\mathrm{bot}}=1$, and no layer is flipped along the out-of-plane direction. For each retained matrix pair, a candidate bilayer structure is constructed using Algorithm~\ref{alg:samba_bilayer_construction} with fixed parameters $d$ and $(\delta_x,\delta_y)=(0,0)$, and the resulting structure $M_{\mathrm{init}}^{\mathrm{bi}}$ is added to the candidate set $\mathcal{M}_{\mathrm{init}}^{\mathrm{bi}}$. After all retained matrix pairs have been processed, duplicate structures in $\mathcal{M_{\mathrm{init}}}^{\mathrm{bi}}$ are removed using RMSD-based geometric matching, with optimal alignment performed via the Kabsch algorithm and atomic correspondence determined using the Hungarian algorithm, yielding the final non-duplicated set $\mathcal{M}_{\mathrm{init}}^{\mathrm{bi}}$ for the given $M^{\mathrm{top}}$ and $M^{\mathrm{bot}}$ monolayers.

\begin{algorithm}[t]
\caption{Initial Bilayer Construction $(\mathrm{ConstructSAMBA})$}
\label{alg:samba_bilayer_construction}
\begin{algorithmic}[1]

\Input 
Monolayers $M^{\mathrm{top}}, M^{\mathrm{bot}}$; 
two-dimensional supercell transformation matrices $C^{\mathrm{top}}, C^{\mathrm{bot}}$; monolayer thicknesses $t^{\mathrm{mono,top}}, t^{\mathrm{mono,bot}}$; vacuum region $v$; interlayer distance $d$

\Output Initial bilayer structure $M_{\mathrm{init}}^{\mathrm{bi}}$

\State Extend the two-dimensional supercell transformation matrices to three-dimensional matrices:
\[
C^{(n)} =
\begin{bmatrix}
c_1^{(n)} & c_2^{(n)} \\
c_3^{(n)} & c_4^{(n)}
\end{bmatrix},
\quad 
S^{(n)} =
\begin{bmatrix}
c_1^{(n)} & c_2^{(n)} & 0 \\
c_3^{(n)} & c_4^{(n)} & 0 \\
0 & 0 & 1
\end{bmatrix},
\quad
n \in \{\mathrm{top},\mathrm{bot}\}.
\]

\State Construct supercells:
\[
M^{(n),\mathrm{sc}} =
(A^{(n),\mathrm{sc}}, P^{(n),\mathrm{sc}}, L^{(n),\mathrm{sc}})
= \mathrm{MakeSupercell}(M^{(n)}, S^{(n)}).
\]

\State Compute common in-plane lattice $L^{\mathrm{sc},\mathrm{common}}$ using Eq.~\eqref{eq:hetdb_common_lattice}

\State Define bilayer lattice vectors:
\[
\mathbf{l}_1^{\mathrm{bi}} = \mathbf{l}_1^{\mathrm{sc},\mathrm{common}}, \quad
\mathbf{l}_2^{\mathrm{bi}} = \mathbf{l}_2^{\mathrm{sc},\mathrm{common}}, \quad
\mathbf{l}_3^{\mathrm{bi}} = (0,0,\,v+d+t^{\mathrm{mono,top}}+t^{\mathrm{mono,bot}})
\]

\State Construct bilayer lattice:
\[
L^{\mathrm{bi}} =
\begin{bmatrix}
\mathbf{l}_1^{\mathrm{bi}},
\mathbf{l}_2^{\mathrm{bi}},
\mathbf{l}_3^{\mathrm{bi}}
\end{bmatrix}^{T}
\]

\State Convert atomic coordinates to fractional coordinates:
\[
F^{(n),\mathrm{sc}} =
P^{(n),\mathrm{sc}} (L^{\mathrm{bi}})^{-1},
\quad n \in \{\mathrm{top},\mathrm{bot}\}
\]

\State Compute the fractional out-of-plane translation:
\[
\delta z =
z_{\max}^{\mathrm{bot}}
+
\frac{d}{\|\mathbf{l}_3^{\mathrm{bi}}\|}
-
z_{\min}^{\mathrm{top}}
\]

\State Define the layer translation vector without in-plane translations:
\[
\boldsymbol{\tau}
=
\begin{bmatrix}
0 & 0 & \delta z
\end{bmatrix}
\]

\State Apply the layer translation to the top layer:
\[
F^{\mathrm{top},\mathrm{sc}}
\leftarrow
\left(
F^{\mathrm{top},\mathrm{sc}}
+
\mathbf{1}\boldsymbol{\tau}
\right)
\bmod 1,
\qquad
\mathbf{1}\in\mathbb{R}^{N^{\mathrm{top},\mathrm{sc}}\times 1}
\]

\State Convert back to Cartesian coordinates:
\[
P^{(n),\mathrm{sc}} =
F^{(n),\mathrm{sc}} L^{\mathrm{bi}},
\quad n \in \{\mathrm{top},\mathrm{bot}\}
\]

\State Combine layers:
\[
P^{\mathrm{bi}} =
P^{\mathrm{top},\mathrm{sc}} \cup P^{\mathrm{bot},\mathrm{sc}}, \quad
A^{\mathrm{bi}} =
A^{\mathrm{top},\mathrm{sc}} \cup A^{\mathrm{bot},\mathrm{sc}}
\]

\State \Return $M_{\mathrm{init}}^{\mathrm{bi}} = (A^{\mathrm{bi}}, P^{\mathrm{bi}}, L^{\mathrm{bi}})$

\end{algorithmic}
\end{algorithm}
The process of constructing an initial bilayer structure in the SAMBA dataset is summarized in Algorithm~\ref{alg:samba_bilayer_construction}. Given two selected two-dimensional supercell transformation matrices, $C^{\mathrm{top}}$ and $C^{\mathrm{bot}}$, for the top and bottom monolayers, the corresponding top and bottom supercells are first constructed. In the current SAMBA construction, the out-of-plane direction is kept unchanged for all materials. Therefore, each integer $2\times2$ supercell transformation matrix $C^{(n)}$ is extended to a $3\times3$ supercell transformation matrix by preserving the out-of-plane direction:
\begin{equation}
C^{(n)} =
\begin{bmatrix}
c_1^{(n)} & c_2^{(n)} \\
c_3^{(n)} & c_4^{(n)}
\end{bmatrix},
\quad 
S^{(n)} =
\begin{bmatrix}
c_1^{(n)} & c_2^{(n)} & 0 \\
c_3^{(n)} & c_4^{(n)} & 0 \\
0 & 0 & 1
\end{bmatrix}, \quad n \in \{\mathrm{top}, \mathrm{bot}\},
\end{equation}
where \(c_i^{(n)} \in \mathbb{Z}\) define the in-plane supercell transformation from \(C^{(n)}\), and the out-of-plane transformation is fixed as $c_5^{(n)}=1$. The top and bottom supercells are then constructed using the same $\mathrm{MakeSupercell}$ function adopted in the HetDB construction procedure. Specifically, \(M^{\mathrm{top},\mathrm{sc}}\) is computed from \(M^{\mathrm{mono,top}}\) and \(S^{\mathrm{top}}\), while \(M^{\mathrm{bot},\mathrm{sc}}\) is computed from \(M^{\mathrm{mono,bot}}\) and \(S^{\mathrm{bot}}\), following Eq.~\eqref{eq:hetdb_make_supercell}.

The common in-plane lattice $L^{\mathrm{sc},\mathrm{common}}$ is computed from $L^{\mathrm{top},\mathrm{sc}}$ and $L^{\mathrm{bot},\mathrm{sc}}$ using Eq.~\eqref{eq:hetdb_common_lattice}. The bilayer out-of-plane length is first computed as $t^{\mathrm{bi}} = v+d+t^{\mathrm{mono,top}}+t^{\mathrm{mono,bot}}$, which includes the top-layer thickness, bottom-layer thickness, interlayer distance, and vacuum spacing. The bilayer lattice is then constructed by using the common lattice for the in-plane directions, with $\mathbf{l}_1^{\mathrm{bi}} = \mathbf{l}_1^{\mathrm{sc},\mathrm{common}}$, $\mathbf{l}_2^{\mathrm{bi}} = \mathbf{l}_2^{\mathrm{sc},\mathrm{common}}$, and $\mathbf{l}_3^{\mathrm{bi}} = (0,0,\,t^{\mathrm{bi}})$. The bilayer lattice matrix is then defined as $L^{\mathrm{bi}} = \begin{bmatrix} \mathbf{l}_1^{\mathrm{bi}}, \mathbf{l}_2^{\mathrm{bi}}, \mathbf{l}_3^{\mathrm{bi}} \end{bmatrix}^{T}$. The atomic positions are converted to fractional coordinates with respect to the bilayer lattice, giving $F^{\mathrm{top},\mathrm{sc}} = P^{\mathrm{top},\mathrm{sc}}(L^{\mathrm{bi}})^{-1}$ and $F^{\mathrm{bot},\mathrm{sc}} = P^{\mathrm{bot},\mathrm{sc}}(L^{\mathrm{bi}})^{-1}$. The fractional out-of-plane translation is then computed from the given initial interlayer distance \(d\), following Eq.~\eqref{eq:bidb_delta_z} used in BiDB. Without in-plane translations, the layer translation vector \(\boldsymbol{\tau}\) is defined using Eq.~\eqref{eq:translation_vector_no_inplane}. The fractional coordinates of the top layer \(F^{\mathrm{top},\mathrm{sc}}\) are then updated using the same layer-translation operation as in Eq.~\eqref{eq:bidb_apply_layer_translation}. Finally, the coordinates are converted back to Cartesian form as $P^{\mathrm{top},\mathrm{sc}} = F^{\mathrm{top},\mathrm{sc}}L^{\mathrm{bi}}$ and $P^{\mathrm{bot},\mathrm{sc}} = F^{\mathrm{bot},\mathrm{sc}}L^{\mathrm{bi}}$. The two layers are then combined as $P^{\mathrm{bi}} = P^{\mathrm{top},\mathrm{sc}} \cup P^{\mathrm{bot},\mathrm{sc}}$ and $A^{\mathrm{bi}} = A^{\mathrm{top},\mathrm{sc}} \cup A^{\mathrm{bot},\mathrm{sc}}$, yielding the initial bilayer structure $M_{\mathrm{init}}^{\mathrm{bi}} = (A^{\mathrm{bi}}, P^{\mathrm{bi}}, L^{\mathrm{bi}})$.

\section{Structural Optimization}
\label{app:structural_optimization}

After the dataset-specific initial bilayer structures $M_{\mathrm{init}}^{\mathrm{bi}}$ are constructed, all initial structures are optimized using the MatterSim-D3 framework. Starting from an initial constructed bilayer structure $M_{\mathrm{init}}^{\mathrm{bi}}$, we first perform a $z$-scan to determine the optimal interlayer distance $d_{\mathrm{opt}}$, as summarized in Algorithm~\ref{alg:zscan}. During the $z$-scan, the in-plane translations are fixed as $\delta x=0$ and $\delta y=0$, and the interlayer distance $d$ is varied using the layer-translation procedure in Algorithm~\ref{alg:layer_translation}. For SAMBA, after $d_{\mathrm{opt}}$ is obtained, an additional $xy$-scan is performed to select the lowest-energy lateral stacking position. Finally, the selected bilayer structure is relaxed using MatterSim-D3 until the maximum atomic force is below the dataset-specific maximum force threshold, yielding the optimized bilayer structure $M^{\mathrm{bi}}$.

For BiDB, we apply a unified $z$-scan procedure to both magnetic and non-magnetic systems. This differs from the original BiDB generation workflow, where non-magnetic systems are optimized using a SciPy-based energy minimization and magnetic systems are optimized using a stepwise $z$-scan with energy normalized by the total number of electrons. In our workflow, the $z$-scan uses the total MatterSim-D3 energy without normalization. The scan starts from $d_{\text{start}}=5.0~\text{\AA}$ and decreases to $d_{\text{end}}=0~\text{\AA}$. The coarse and fine step sizes are set to $s_{\text{coarse}}=0.25~\text{\AA}$ and $s_{\text{fine}}=0.05~\text{\AA}$, respectively. The coarse scan first identifies an approximate minimum-energy distance. Starting from this distance, the fine scan further decreases the interlayer distance and is terminated once the energy increases relative to the previous step. The previous distance is selected as $d_{\mathrm{opt}}$. The final BFGS relaxation is performed until the maximum atomic force is below $0.01~\mathrm{eV/\text{\AA}}$, producing $M^{\mathrm{bi}}$.

For HetDB, each initial heterogeneous bilayer structure $M^{\mathrm{bi}}$ is optimized using the same MatterSim-D3-based $z$-scan and BFGS relaxation workflow. The $z$-scan starts from $d_{\text{start}}=10.0~\text{\AA}$ and decreases to $d_{\text{end}}=0~\text{\AA}$, with coarse and fine step sizes of $s_{\text{coarse}}=0.25~\text{\AA}$ and $s_{\text{fine}}=0.05~\text{\AA}$, respectively. The coarse scan is used to locate the approximate minimum-energy distance, and the fine scan is then performed from this distance with step size $s_{\text{fine}}$. The fine scan stops when the energy increases relative to the previous step, and the previous distance is selected as the optimal interlayer distance $d_{\mathrm{opt}}$. The final BFGS relaxation is performed until the maximum atomic force is below $0.05~\mathrm{eV/\text{\AA}}$, producing $M^{\mathrm{bi}}$.

For SAMBA, each non-duplicated initial bilayer structure $M_{\mathrm{init}}^{\mathrm{bi}}\in\mathcal{M}_{\mathrm{init}}^{\mathrm{bi}}$ is optimized separately. The interlayer distance is first optimized using the $z$-scan procedure in Algorithm~\ref{alg:zscan}, where the layer-translation procedure in Algorithm~\ref{alg:layer_translation} is applied with $\delta x=0$ and $\delta y=0$. The $z$-scan starts from $d_{\text{start}}=5.0~\text{\AA}$ and decreases to $d_{\text{end}}=0~\text{\AA}$, with coarse and fine step sizes $s_{\text{coarse}}=0.5~\text{\AA}$ and $s_{\text{fine}}=0.05~\text{\AA}$, respectively. The coarse scan first locates an approximate minimum-energy distance. The fine scan then starts from this distance and decreases $d$ until the MatterSim-D3 total energy increases relative to the previous step, after which the previous distance is selected as $d_{\mathrm{opt}}$. After obtaining $d_{\mathrm{opt}}$, an additional $xy$-scan is performed to optimize the lateral stacking registry. This scan also uses the layer-translation procedure in Algorithm~\ref{alg:layer_translation}, but with the interlayer distance fixed at $d_{\mathrm{opt}}$. Specifically, a $9\times9$ grid of in-plane translations $(\delta x,\delta y)$, uniformly sampled from $0$ to $5/6$ along both in-plane directions, is evaluated. The translation pair with the lowest MatterSim-D3 total energy is selected as $(\delta x,\delta y)_{\mathrm{opt}}$. The final BFGS relaxation is performed until the maximum atomic force is below $0.01~\mathrm{eV/\text{\AA}}$, producing $M^{\mathrm{bi}}$.

\label{app:layer_translation}
\begin{algorithm}[t]
\caption{Layer Translation $(\mathrm{LayerTranslation})$}
\label{alg:layer_translation}
\begin{algorithmic}[1]
\Input Bilayer material $M^{\mathrm{bi}} = (A^{\mathrm{bi}}, P^{\mathrm{bi}}, L^{\mathrm{bi}})$, dataset type $\mathcal{D}$, in-plane translations $(\delta x,\delta y)$, interlayer distance $d$
\Output Translated bilayer material $M^{\mathrm{bi}}_{\mathrm{translated}}$

\State Compute the layer-separation cutoff $z_{\mathrm{cut}}$ using Eq.~\eqref{eq:layer_translation_z_cut}

\State Split the bilayer positions using $z_{\mathrm{cut}}$:
\[
P^{\mathrm{bot}}
=
\left\{
p_i^{\mathrm{bi}} \in P^{\mathrm{bi}}
\,\middle|\,
p_i^{3,\mathrm{bi}} < z_{\mathrm{cut}}
\right\},
\qquad
P^{\mathrm{top}}
=
\left\{
p_i^{\mathrm{bi}} \in P^{\mathrm{bi}}
\,\middle|\,
p_i^{3,\mathrm{bi}} \ge z_{\mathrm{cut}}
\right\}
\]

\State Convert the positions to fractional coordinates:
\[
F^{\mathrm{bot}}
=
P^{\mathrm{bot}}
(L^{\mathrm{bi}})^{-1},
\qquad
F^{\mathrm{top}}
=
P^{\mathrm{top}}
(L^{\mathrm{bi}})^{-1}
\]

\State Compute the required fractional out-of-plane layer translation $\delta z$:
\[
\delta z =
\begin{cases}
z_{\max}^{\mathrm{bot}}
+
\dfrac{d}{\|\mathbf{l}_3^{\mathrm{bi}}\|}
-
z_{\min}^{\mathrm{top}},
&
\mathcal{D}\in\{\mathrm{BiDB},\mathrm{SAMBA}\},
\\[8pt]
\dfrac{
d -
\left(
z_{\mathrm{mean}}^{\mathrm{top}}
-
z_{\mathrm{mean}}^{\mathrm{bot}}
\right)
\|\mathbf{l}_3^{\mathrm{bi}}\|
}
{\|\mathbf{l}_3^{\mathrm{bi}}\|},
&
\mathcal{D}=\mathrm{HetDB}.
\end{cases}
\]

\State Define the layer translation vector:
\[
\boldsymbol{\tau}
=
\begin{bmatrix}
\delta x & \delta y & \delta z
\end{bmatrix}
\]

\State Apply the layer translation to the top layer:
\[
F^{\mathrm{top}}_{\mathrm{translated}}
\leftarrow
\left(
F^{\mathrm{top}}
+
\mathbf{1}\boldsymbol{\tau}
\right)
\bmod 1,
\qquad
\mathbf{1}\in\mathbb{R}^{N_{\mathrm{top}}\times 1}
\]

\State Convert the translated top layer back to Cartesian coordinates:
\[
P^{\mathrm{top}}_{\mathrm{translated}}
=
F^{\mathrm{top}}_{\mathrm{translated}}L^{\mathrm{bi}}
\]

\State Construct the translated bilayer:
\[
P^{\mathrm{bi}}_{\mathrm{translated}}
=
P^{\mathrm{bot}}
\cup
P^{\mathrm{top}}_{\mathrm{translated}}
\]

\[
M^{\mathrm{bi}}_{\mathrm{translated}}
=
(A^{\mathrm{bi}},P^{\mathrm{bi}}_{\mathrm{translated}},L^{\mathrm{bi}})
\]

\State \Return $M^{\mathrm{bi}}_{\mathrm{translated}}$
\end{algorithmic}
\end{algorithm}

The layer translation workflow follows Algorithm~\ref{alg:layer_translation}. This procedure can be used in both the $z$-scan and the $xy$-scan to update the atomic coordinates of a bilayer structure. Given a bilayer material $M^{\mathrm{bi}} = (A^{\mathrm{bi}}, P^{\mathrm{bi}}, L^{\mathrm{bi}})$, in-plane translations $(\delta x,\delta y)$, and an interlayer distance $d$, the goal is to translate $M^{\mathrm{bi}}$ into $M^{\mathrm{bi}}_{\mathrm{translated}}$. In the $z$-scan, the in-plane translations are fixed as $\delta x = 0$ and $\delta y = 0$, while $d$ is varied to find the optimal interlayer distance $d_{\mathrm{opt}}$. After $d_{\mathrm{opt}}$ is obtained, $d$ is fixed to $d_{\mathrm{opt}}$, and the same layer-translation procedure can be used for the $xy$-scan by varying $(\delta x,\delta y)$.
In the layer-translation algorithm, the Cartesian $z$-coordinate threshold $z_{\mathrm{cut}}$ used to separate the bottom and top layers is first defined as the midpoint between the lowest and highest atomic Cartesian $z$-coordinates of the bilayer:
\begin{equation}
\label{eq:layer_translation_z_cut}
z_{\mathrm{cut}}
=
\frac{
\min_i p_i^{3,\mathrm{bi}}
+
\max_i p_i^{3,\mathrm{bi}}
}{2},
\end{equation}
where $p_i^{3,\mathrm{bi}}$ denotes the Cartesian $z$-coordinate of atom $i$ in $P^{\mathrm{bi}}$. The bilayer atomic positions are then split into the bottom and top layers using \(z_{\mathrm{cut}}\):
\begin{equation}
\label{eq:split_bilayer_layers}
P^{\mathrm{bot}}
=
\left\{
p_i^{\mathrm{bi}} \in P^{\mathrm{bi}}
\,\middle|\,
p_i^{3,\mathrm{bi}} < z_{\mathrm{cut}}
\right\},
\qquad
P^{\mathrm{top}}
=
\left\{
p_i^{\mathrm{bi}} \in P^{\mathrm{bi}}
\,\middle|\,
p_i^{3,\mathrm{bi}} \ge z_{\mathrm{cut}}
\right\}.
\end{equation}
Both layers are then converted into fractional coordinates using the bilayer lattice \(L^{\mathrm{bi}}\), giving \(F^{\mathrm{bot}} = P^{\mathrm{bot}}(L^{\mathrm{bi}})^{-1}\) and \(F^{\mathrm{top}} = P^{\mathrm{top}}(L^{\mathrm{bi}})^{-1}\). To impose the target interlayer distance $d$, the fractional out-of-plane translation $\delta z$ is computed according to the dataset construction protocol:
\begin{equation}
\delta z =
\begin{cases}
z_{\max}^{\mathrm{bot}}
+
\dfrac{d}{\|\mathbf{l}_3^{\mathrm{bi}}\|}
-
z_{\min}^{\mathrm{top}},
&
\mathcal{D}\in\{\mathrm{BiDB},\mathrm{SAMBA}\},
\\[8pt]
\dfrac{
d -
\left(
z_{\mathrm{mean}}^{\mathrm{top}}
-
z_{\mathrm{mean}}^{\mathrm{bot}}
\right)
\|\mathbf{l}_3^{\mathrm{bi}}\|
}
{\|\mathbf{l}_3^{\mathrm{bi}}\|},
&
\mathcal{D}=\mathrm{HetDB}.
\end{cases}
\end{equation}
where $z_{\min}^{\mathrm{top}}$ denotes the minimum fractional $z$-coordinate of the top layer, $z_{\max}^{\mathrm{bot}}$ denotes the maximum fractional $z$-coordinate of the bottom layer, and $z_{\mathrm{mean}}^{\mathrm{top}}$ and $z_{\mathrm{mean}}^{\mathrm{bot}}$ denote the mean fractional $z$-coordinates of the top and bottom layers, respectively. The layer translation vector \(\boldsymbol{\tau}\) is defined using Eq.~\eqref{eq:bidb_layer_translation_vector}. The bottom layer is kept fixed, while the layer translation is applied only to the top layer following Eq.~\eqref{eq:bidb_apply_layer_translation}. The translated top-layer fractional coordinates are converted back to Cartesian coordinates using the same bilayer lattice as $P^{\mathrm{top}}_{\mathrm{translated}} = F^{\mathrm{top}}_{\mathrm{translated}}L^{\mathrm{bi}}$. Finally, the translated bilayer positions are constructed as $P^{\mathrm{bi}}_{\mathrm{translated}} = P^{\mathrm{bot}} \cup P^{\mathrm{top}}_{\mathrm{translated}}$, and the translated bilayer material is given by $M^{\mathrm{bi}}_{\mathrm{translated}} = (A^{\mathrm{bi}},P^{\mathrm{bi}}_{\mathrm{translated}},L^{\mathrm{bi}})$.

\begin{algorithm}[t]
\caption{z-scan $(\mathrm{z-scan})$}
\label{alg:zscan}
\begin{algorithmic}[1]

\State \textbf{Input:} Bilayer material $M^{\mathrm{bi}} = (A^{\mathrm{bi}}, P^{\mathrm{bi}}, L^{\mathrm{bi}})$, dataset type $\mathcal{D}$, search range $d_{\text{start}}$, $d_{\text{end}}$, coarse step size $s_{\text{coarse}}$, fine step size $s_{\text{fine}}$, MLIP model $\mathrm{MatterSim\mbox{-}D3}$
\State \textbf{Output:} Optimal interlayer distance $d_{\mathrm{opt}}$

\State \textbf{Stage 1: Coarse scan}

\State $d_{\text{current}} \leftarrow d_{\text{start}}$
\State $d_{\mathrm{opt}} \leftarrow d_{\text{current}}$
\State $E(d_{\mathrm{opt}}) \leftarrow +\infty$

\While{$d_{\text{current}} > d_{\text{end}}$}
    \State $M^{\mathrm{bi}}_{\mathrm{translated}} \leftarrow \mathrm{LayerTranslation}(M^{\mathrm{bi}}, \mathcal{D}, 0, 0, d_{\text{current}})$
    \State $E(d_{\text{current}}) \leftarrow \mathrm{MatterSim\mbox{-}D3}(M^{\mathrm{bi}}_{\mathrm{translated}})$
    \If{$E(d_{\text{current}}) < E(d_{\mathrm{opt}})$}
        \State $E(d_{\mathrm{opt}}) \leftarrow E(d_{\text{current}})$
        \State $d_{\mathrm{opt}} \leftarrow d_{\text{current}}$
    \EndIf
    \State $d_{\text{current}} \leftarrow d_{\text{current}} - s_{\text{coarse}}$
\EndWhile

\State \textbf{Stage 2: Fine scan}

\State $d_{\text{current}} \leftarrow d_{\mathrm{opt}}$
\State $d_{\text{previous}} \leftarrow d_{\text{current}}$
\State $E(d_{\text{previous}}) \leftarrow E(d_{\mathrm{opt}})$

\While{True}
    \State $d_{\text{current}} \leftarrow d_{\text{current}} - s_{\text{fine}}$
    \State $M^{\mathrm{bi}}_{\mathrm{translated}} \leftarrow \mathrm{LayerTranslation}(M^{\mathrm{bi}}, \mathcal{D}, 0, 0, d_{\text{current}})$
    \State $E(d_{\text{current}}) \leftarrow \mathrm{MatterSim\mbox{-}D3}(M^{\mathrm{bi}}_{\mathrm{translated}})$
    \If{$E(d_{\text{current}}) > E(d_{\text{previous}})$}
        \State \textbf{break}
    \EndIf
    \State $d_{\text{previous}} \leftarrow d_{\text{current}}$
    \State $E(d_{\text{previous}}) \leftarrow E(d_{\text{current}})$
\EndWhile

\State $d_{\mathrm{opt}} \leftarrow d_{\text{previous}}$

\State \Return $d_{\mathrm{opt}}$

\end{algorithmic}
\end{algorithm}

The $z$-scan procedure implemented in this work is summarized in Algorithm~\ref{alg:zscan}. Because the original $z$-scan implementations are dataset-specific and may include details that are not fully specified in the published descriptions, our implementation may not exactly match the original code used to generate each dataset. Starting from an initial bilayer structure $M^{\mathrm{bi}}$, the algorithm searches for the optimal interlayer distance $d_{\mathrm{opt}}$ by evaluating the $\mathrm{MatterSim\mbox{-}D3}$ total energy over a sequence of candidate distances. For each candidate distance $d_{\text{current}}$, the layer-translation procedure in Algorithm~\ref{alg:layer_translation} is applied with fixed in-plane translations $\delta x=0$ and $\delta y=0$ to update the separation between the two layers. The scan is performed in two stages. First, a coarse scan from $d_{\text{start}}$ to $d_{\text{end}}$ with step size $s_{\text{coarse}}$ is used to locate the approximate minimum-energy distance. Then, a fine scan starts from this distance and decreases $d$ with step size $s_{\text{fine}}$. The fine scan continues as long as the energy does not increase relative to the previous step. Once the energy increases, the scan is terminated, and the previous distance, corresponding to the lowest energy encountered in the fine scan, is selected as the optimal interlayer distance $d_{\mathrm{opt}}$.

\section{BDIP-Net For Bilayer Property Prediction}
\subsection{BDIP-Net Algorithm}
\label{app:BDIP-Net}

\begin{algorithm}[t]
\caption{BDIP-Net Algorithm for Bilayer Property Prediction}
\label{alg:BDIP-Net}
\begin{algorithmic}[1]
\Input 
Structurally optimized bilayer material 
$M^{\mathrm{bi}}=(A^{\mathrm{bi}},P^{\mathrm{bi}},L^{\mathrm{bi}})$;
initial cutoffs $R^{\mathrm{intra}},R^{\mathrm{inter}}$;
required neighbor counts $N^{\mathrm{intra}},N^{\mathrm{inter}}$;
scaling coefficients 
$\epsilon^{\mathrm{Coulomb}},\epsilon^{\mathrm{London}}$;
ground-truth property $Y$
\Output Predicted bilayer property $\hat{Y}$

\State $\mathcal{E}^{\mathrm{intra}} \gets 
\mathrm{AdaptiveLayerNeighborSearch}
\left(
M^{\mathrm{bi}}, 
R^{\mathrm{intra}}, 
N^{\mathrm{intra}}, 
\mathrm{intra}
\right)$

\State $\mathcal{E}^{\mathrm{inter}} \gets 
\mathrm{AdaptiveLayerNeighborSearch}
\left(
M^{\mathrm{bi}},
R^{\mathrm{inter}}, 
N^{\mathrm{inter}}, 
\mathrm{inter}
\right)$

\For{each edge $(i,j) \in \mathcal{E}^{\mathrm{intra}}$}
    \State $e_{ij}^{\mathrm{intra}}
    \gets
    \mathrm{MLP}_{e}^{\mathrm{intra}}
    \left(
    \mathrm{RBF}^{\mathrm{Coulomb}}
    \left(
    -\epsilon^{\mathrm{Coulomb}}/d_{ij}
    \right)
    \right)$
\EndFor

\For{each edge $(i,j) \in \mathcal{E}^{\mathrm{inter}}$}
    \State $e_{ij}^{\mathrm{inter}}
    \gets
    \mathrm{MLP}_{e}^{\mathrm{inter}}
    \left(
    \mathrm{RBF}^{\mathrm{London}}
    \left(
    -\epsilon^{\mathrm{London}}/(d_{ij})^{6}
    \right)
    \right)$
\EndFor
\State $\mathcal{V}^{(0)} 
\gets 
\mathrm{Linear}_{\mathcal{V}}
\left(
A^{\mathrm{bi}}
\right)$

\For{$t=0$ \textbf{to} $T-1$}
    \For{each atom $i \in \mathcal{V}$}
        \For{each interaction type 
        $\eta \in \{\mathrm{intra},\mathrm{inter}\}$}
            \State $m_i^{(t),\eta} \gets \mathbf{0}$

            \For{each edge $(i,j) \in \mathcal{E}^{\eta}_i$}
                \State $w_{ij}^{(t),\eta}
                \gets
                v_i^{(t)} \oplus v_j^{(t)} \oplus e_{ij}^{\eta}$

                \State $z_{ij}^{(t),\eta}
                \gets
                \mathrm{MLP}_{z}^{(t),\eta}
                \left(
                w_{ij}^{(t),\eta}
                \right)$

                \State $\alpha_{ij}^{(t),\eta}
                \gets
                \operatorname{softmax}_{j \in \mathcal{N}^{\eta}_{(i)}}
                \left(
                \mathrm{MLP}_{\alpha}^{(t),\eta}
                \left(
                z_{ij}^{(t),\eta}
                \right)
                \right)$

                \State $m_i^{(t),\eta}
                \gets
                m_i^{(t),\eta}
                +
                \alpha_{ij}^{(t),\eta}
                z_{ij}^{(t),\eta}$
            \EndFor
        \EndFor

        \State
        $\begin{bmatrix}
        \beta_i^{(t),\mathrm{intra}}\\
        \beta_i^{(t),\mathrm{inter}}
        \end{bmatrix}
        \gets
        \operatorname{softmax}
        \left(
        \mathrm{MLP}_{\beta}^{(t)}
        \left(
        \begin{bmatrix}
        m_i^{(t),\mathrm{intra}}\\
        m_i^{(t),\mathrm{inter}}
        \end{bmatrix}
        \right)
        \right)$

        \State $v_i^{(t+1)}
        \gets
        v_i^{(t)}
        +
        \beta_i^{(t),\mathrm{intra}}
        m_i^{(t),\mathrm{intra}}
        +
        \beta_i^{(t),\mathrm{inter}}
        m_i^{(t),\mathrm{inter}}$
    \EndFor
\EndFor

\State $Z_{\mathcal{G}}
\gets
\frac{1}{|\mathcal{V}|}
\sum_{i \in \mathcal{V}}
v_i^{(T)}$

\State $\hat{Y}
\gets
\mathrm{MLP}_{\hat{Y}}
\left(
Z_{\mathcal{G}}
\right)$

\State $\mathcal{L} \gets |\hat{Y}-Y|$

\State Update all trainable parameters by minimizing $\mathcal{L}$

\State \Return $\hat{Y}$
\end{algorithmic}
\end{algorithm}

Algorithm~\ref{alg:BDIP-Net} summarizes the complete training and prediction procedure of BDIP-Net for bilayer material property prediction. The algorithm takes as input a structurally optimized bilayer material $M^{\mathrm{bi}}=(A^{\mathrm{bi}},P^{\mathrm{bi}},L^{\mathrm{bi}})$, the initial intra-layer and inter-layer cutoff radii $R^{\mathrm{intra}}$ and $R^{\mathrm{inter}}$, the required neighbor counts $N^{\mathrm{intra}}$ and $N^{\mathrm{inter}}$, the potential scaling coefficients $\epsilon^{\mathrm{Coulomb}}$ and $\epsilon^{\mathrm{London}}$, and the ground-truth property $Y$ during training. The output is the predicted bilayer property $\hat{Y}$. Lines~1--2 construct the interaction-specific edge sets $\mathcal{E}^{\mathrm{intra}}$ and $\mathcal{E}^{\mathrm{inter}}$ using the adaptive layer neighbor search procedure. Lines~3--8 compute the intra-layer and inter-layer edge embeddings $e_{ij}^{\mathrm{intra}}$ and $e_{ij}^{\mathrm{inter}}$ from the corresponding Coulomb-based and London-based potentials. Line~9 projects the atomic features into the initial node embeddings $\mathcal{V}^{(0)}=\{v_i^{(0)}\}_{i=1}^{N}$. Lines~10--24 perform $T$ message-passing convolution layers. Within each convolution layer, the message-passing operations are performed separately for each interaction type $\eta\in\{\mathrm{intra},\mathrm{inter}\}$. Lines~15--16 construct the interaction-specific representation $w_{ij}^{(t),\eta}$ from $v_i^{(t)}$, $v_j^{(t)}$, and $e_{ij}^{\eta}$, and transform $w_{ij}^{(t),\eta}$ into the corresponding edge message $z_{ij}^{(t),\eta}$. Lines~17--18 compute the interaction-specific edge attention coefficient $\alpha_{ij}^{(t),\eta}$ from $z_{ij}^{(t),\eta}$, use $\alpha_{ij}^{(t),\eta}$ to weight the corresponding edge message $z_{ij}^{(t),\eta}$, and aggregate the weighted edge messages over neighbors of interaction type $\eta$ to obtain the interaction-specific node message $m_i^{(t),\eta}$. Consequently, each node obtains an intra-layer message $m_i^{(t),\mathrm{intra}}$ and an inter-layer message $m_i^{(t),\mathrm{inter}}$. Lines~21--22 compute the interaction attention weights $\beta_i^{(t),\mathrm{intra}}$ and $\beta_i^{(t),\mathrm{inter}}$ from $m_i^{(t),\mathrm{intra}}$ and $m_i^{(t),\mathrm{inter}}$, respectively, use them to weight and fuse the two interaction-specific node messages, and add the fused message to the residual node representation $v_i^{(t)}$ to obtain the updated node embedding $v_i^{(t+1)}$. After the $T$ message-passing convolution layers, Line~25 performs mean pooling over the final node embeddings $\{v_i^{(T)}\}_{i\in\mathcal{V}}$ to obtain the graph representation $Z_{\mathcal G}$. Line~26 transforms $Z_{\mathcal G}$ into the predicted bilayer property $\hat{Y}$. During training, Lines~27--28 compute the MAE loss $\mathcal{L}$ from $\hat{Y}$ and $Y$ and update all trainable parameters through backpropagation. During inference, Line~29 returns the predicted bilayer property $\hat{Y}$ after the forward pass.

\subsection{Interaction-Specific Radial Basis Function Expansions}
\label{app:rbf}

BDIP-Net uses separate radial basis function (RBF) expansions for the
Coulomb and London dispersion potential values. Each expansion maps a scalar
potential value to a $d_e$-dimensional feature vector.
The RBF centers are defined over the potential ranges induced by the distance
interval $d\in[d_{\min},d_{\max}]$, where $d_{\min}=1$ and $d_{\max}=8$ in
our experiments. For the Coulomb potential, the RBF centers are uniformly
placed over
\[
\left[
-\frac{\epsilon^{\mathrm{Coulomb}}}{d_{\max}},
-\frac{\epsilon^{\mathrm{Coulomb}}}{d_{\min}}
\right].
\]
For the London dispersion potential, the RBF centers are uniformly placed over
\[
\left[
-\frac{\epsilon^{\mathrm{London}}}{(d_{\max})^{6}},
-\frac{\epsilon^{\mathrm{London}}}{(d_{\min})^{6}}
\right].
\]

The resulting RBF-expanded features are mapped to $d_h$-dimensional edge
embeddings by the independently parameterized networks
$\mathrm{MLP}_{e}^{\mathrm{intra}}$ and
$\mathrm{MLP}_{e}^{\mathrm{inter}}$.
\subsection{Adaptive Layer Neighbor Search}
\label{app:layer_neighbor_search}

\begin{algorithm}[t]
\caption{Adaptive Layer Neighbor Search}
\label{alg:adaptive_layer_neighbor}
\begin{algorithmic}[1]

\Input
Structurally optimized bilayer material
$M^{\mathrm{bi}}=(A,P,L)$;
initial cutoff radius $R$;
required number of neighbors $N$;
interaction type
$\eta \in \{\mathrm{intra},\mathrm{inter}\}$

\Output
Interaction-specific edge set $\mathcal{E}^{\eta}$

\State Compute the layer-separation cutoff
$z^{\mathrm{cut}}$ using
Eq.~\eqref{eq:layer_translation_z_cut}

\For{each atom $i \in A$}
    \If{$z_i > z^{\mathrm{cut}}$}
        \State $\ell_i \gets 1$
    \Else
        \State $\ell_i \gets 0$
    \EndIf
\EndFor

\State $\mathit{enough} \gets \texttt{False}$

\While{not $\mathit{enough}$}

    \State Compute
    $\mathcal{N}_i
    \gets
    \mathrm{FindNeighbors}(M^{\mathrm{bi}},R)$
    for each atom $i \in A$

    \If{$\eta=\mathrm{intra}$}
        \For{each atom $i \in A$}
            \State
            $\mathcal{N}_i
            \gets
            \{(j,\mathbf{k},d_{ij})\in\mathcal{N}_i
            \mid \ell_i=\ell_j\}$
        \EndFor
    \ElsIf{$\eta=\mathrm{inter}$}
        \For{each atom $i \in A$}
            \State
            $\mathcal{N}_i
            \gets
            \{(j,\mathbf{k},d_{ij})\in\mathcal{N}_i
            \mid \ell_i\neq\ell_j\}$
        \EndFor
    \EndIf

    \State $\mathit{enough} \gets \texttt{True}$

    \For{each atom $i \in A$}
        \State Sort $\mathcal{N}_i$ by ascending $d_{ij}$

        \If{$|\mathcal{N}_i|\ge N$}
            \State
            $d_i^{(N)}
            \gets
            \text{distance to the $N$-th neighbor in }\mathcal{N}_i$

            \State
            $\mathcal{N}_i^{\mathrm{filtered}}
            \gets
            \{(j,\mathbf{k},d_{ij})\in\mathcal{N}_i
            \mid d_{ij}\le d_i^{(N)}\}$
        \Else
            \State
            $\mathcal{N}_i^{\mathrm{filtered}}
            \gets
            \mathcal{N}_i$

            \State
            $\mathit{enough}
            \gets
            \texttt{False}$
        \EndIf
    \EndFor

    \If{not $\mathit{enough}$}
        \State $R \gets 2R$
    \EndIf

\EndWhile

\State $\mathcal{E}^{\eta}\gets\emptyset$

\For{each atom $i \in A$}
    \For{each
    $(j,\mathbf{k},d_{ij})
    \in
    \mathcal{N}_i^{\mathrm{filtered}}$}
        \State
        Add edge $(i,j)_{\mathbf{k}}$
        to $\mathcal{E}^{\eta}$
    \EndFor
\EndFor

\State \Return $\mathcal{E}^{\eta}$

\end{algorithmic}
\end{algorithm}

Algorithm~\ref{alg:adaptive_layer_neighbor} describes an adaptive neighbor search for a specified interaction type applied to a structurally optimized bilayer material $M^{\mathrm{bi}}=(A,P,L)$. The algorithm first computes the layer-separation cutoff $z^{\mathrm{cut}}$ using Eq.~\eqref{eq:layer_translation_z_cut} and assigns each atom to either the top or bottom layer. Atoms with $z_i > z^{\mathrm{cut}}$ are assigned to the top layer, whereas atoms with $z_i \le z^{\mathrm{cut}}$ are assigned to the bottom layer. It then performs a neighbor search with an initial cutoff radius $R$ and filters the resulting neighbors according to the selected interaction type $\eta$. For $\eta=\mathrm{intra}$, only neighbors satisfying $\ell_i=\ell_j$ are retained, whereas for $\eta=\mathrm{inter}$, only neighbors satisfying $\ell_i\neq\ell_j$ are retained. The filtered neighbor list of each atom is sorted by distance. If an atom has at least $N$ valid neighbors, the distance to its $N$-th nearest valid neighbor is used as an adaptive cutoff, and all valid neighbors within this distance are retained. Consequently, the complete neighbor shell at the $N$-th distance is preserved, including any additional neighbors located at the same distance as the $N$-th neighbor. Otherwise, the cutoff radius is doubled and the neighbor search is repeated until every atom has at least $N$ valid neighbors. The resulting filtered neighbor lists are then used to construct the interaction-specific edge set $\mathcal{E}^{\eta}$.

\section{Additional Experimental Results}

\subsection{Datasets}

We conduct our evaluation on BiDB, HetDB, and SAMBA because they provide complementary coverage of aligned homobilayers, heterobilayers, and twisted bilayer systems, together forming a diverse benchmark for evaluating our MatterSim-D3-based structural optimization workflow and the interaction-aware graph learning capability of BDIP-Net.

\noindent\textbf{Van der Waals Bilayer Database.}
We follow Bimat-ML~\cite{vuong2026bimatml} and use the same processed BiDB dataset. The dataset contains homogeneous bilayers derived from monolayers in the C2DB database. After removing 250 bilayers associated with 10 monolayers lacking CIF files, the processed dataset contains 10,899 valid bilayer structures. We use the bandgap as the target prediction property and exclude bilayers without valid bandgap values, resulting in a final dataset of 6,683 bilayer materials formed from 940 unique monolayers.

\noindent\textbf{Van der Waals 2D Heterostructure Database.}
We follow Bimat-ML~\cite{vuong2026bimatml} and use the same HetDB dataset. The dataset comprises 336 heterogeneous bilayer materials constructed from 38 distinct monolayers in the C2DB database. Each bilayer is formed by stacking two different monolayers, and no duplicate bilayers are generated from the same monolayer pair under different stacking configurations. Bandgap values are available for all 336 bilayers and are used as the target property.

\noindent\textbf{Simulation and Automated Methods for Bilayer Analysis.}
We use the SAMBA dataset, which contains more than 18,000 twisted bilayer structures generated from 63 monolayers in the C2DB database. Due to the high computational cost of DFT calculations, only 980 bilayer structures have undergone DFT structural optimization and property calculations, including 144 homogeneous bilayer materials and 836 heterogeneous bilayer materials. Bandgap values are available for all 980 DFT-calculated bilayers and are used as the target property.


\begin{table}[t]
\centering
\caption{Bilayer types and bandgap relationships in the BiDB, HetDB, and SAMBA datasets.}
\label{tab:gap_relation}
\resizebox{\columnwidth}{!}{%
\begin{tabular}{lccccccc}
\toprule
\multirow{2}{*}{\textbf{Dataset}} &
\multirow{2}{*}{\textbf{\makecell{Bilayer\\Type}}} &
\multirow{2}{*}{\textbf{\makecell{Number of\\Bilayers}}} &
\multirow{2}{*}{\textbf{\makecell{Mean / Min / Max\\Bandgap (eV)}}} &
\multicolumn{2}{c}{\textbf{Monolayer Bandgap Range}} &
\multicolumn{2}{c}{\textbf{Close to Monolayer Bandgaps}} \\
\cmidrule(lr){5-6}
\cmidrule(lr){7-8}
&
&
&
&
\textbf{Within} &
\textbf{Outside} &
\textbf{Close} &
\textbf{Neither} \\
\midrule

BiDB
& Homo
& 6683
& 0.93 / 0.00 / 5.76
& N/A
& N/A
& 4723
& 1960 \\
\midrule
HetDB
& Hetero
& 336
& 0.76 / 0.00 / 3.35
& 76
& 260
& 149
& 187 \\
\midrule
\multirow{2}{*}{SAMBA}
& Homo
& 146
& 0.66 / 0.00 / 4.45
& N/A
& N/A
& 75
& 71 \\

& Hetero
& 834
& 0.28 / 0.00 / 2.01
& 435
& 399
& 607
& 227 \\

\bottomrule
\end{tabular}%
}
\end{table}

\noindent\textbf{Dataset statistics.}
Table~\ref{tab:gap_relation} 
summarizes the composition and bandgap distributions of the three datasets. BiDB contains 6,683 homobilayers with a mean bandgap of \(0.93\) eV, while HetDB contains 336 heterobilayers with a mean bandgap of \(0.76\) eV. SAMBA contains 146 homobilayers and 834 heterobilayers, with mean bandgaps of \(0.66\) and \(0.28\) eV, respectively. The SAMBA heterobilayers are therefore concentrated at lower bandgap values, whereas the homobilayers cover a wider bandgap range. In this statistical analysis, we use a threshold of \(0.2\) eV and consider a bilayer bandgap to be close to one of the two monolayer bandgaps forming the bilayer when their absolute difference does not exceed this threshold. Under this criterion, 4,723 BiDB homobilayers are close to their corresponding monolayer bandgaps, while 1,960 are not. For heterobilayers, the monolayer bandgap range is defined as the interval between the smaller and larger monolayer bandgaps. In HetDB, 76 bilayers lie within this range and 260 lie outside it. Among all HetDB samples, 149 are close to the smaller monolayer bandgap, while none are close to the larger one. In SAMBA, 435 heterobilayers lie within the monolayer bandgap range and 399 lie outside it. Among the 605 samples close to a monolayer bandgap, 603 are close to the smaller bandgap and only 2 are close to the larger one. These results show that, after stacking, when a heterobilayer bandgap remains close to one of its monolayer bandgaps, it tends to be closer to the smaller monolayer bandgap. For the SAMBA homobilayers, 75 samples are close to the corresponding monolayer bandgap, while 71 are not. Overall, the differences in dataset size, bilayer type, bandgap distribution, and bilayer-monolayer bandgap relationships introduce substantial distributional variation and make cross-dataset prediction more challenging.

\subsection{Running Time}
\label{app:runningtime}

\noindent\textbf{Model Training and Inference Runtime.}
\begin{figure}[H]
    \centering
    \includegraphics[width=1\linewidth]{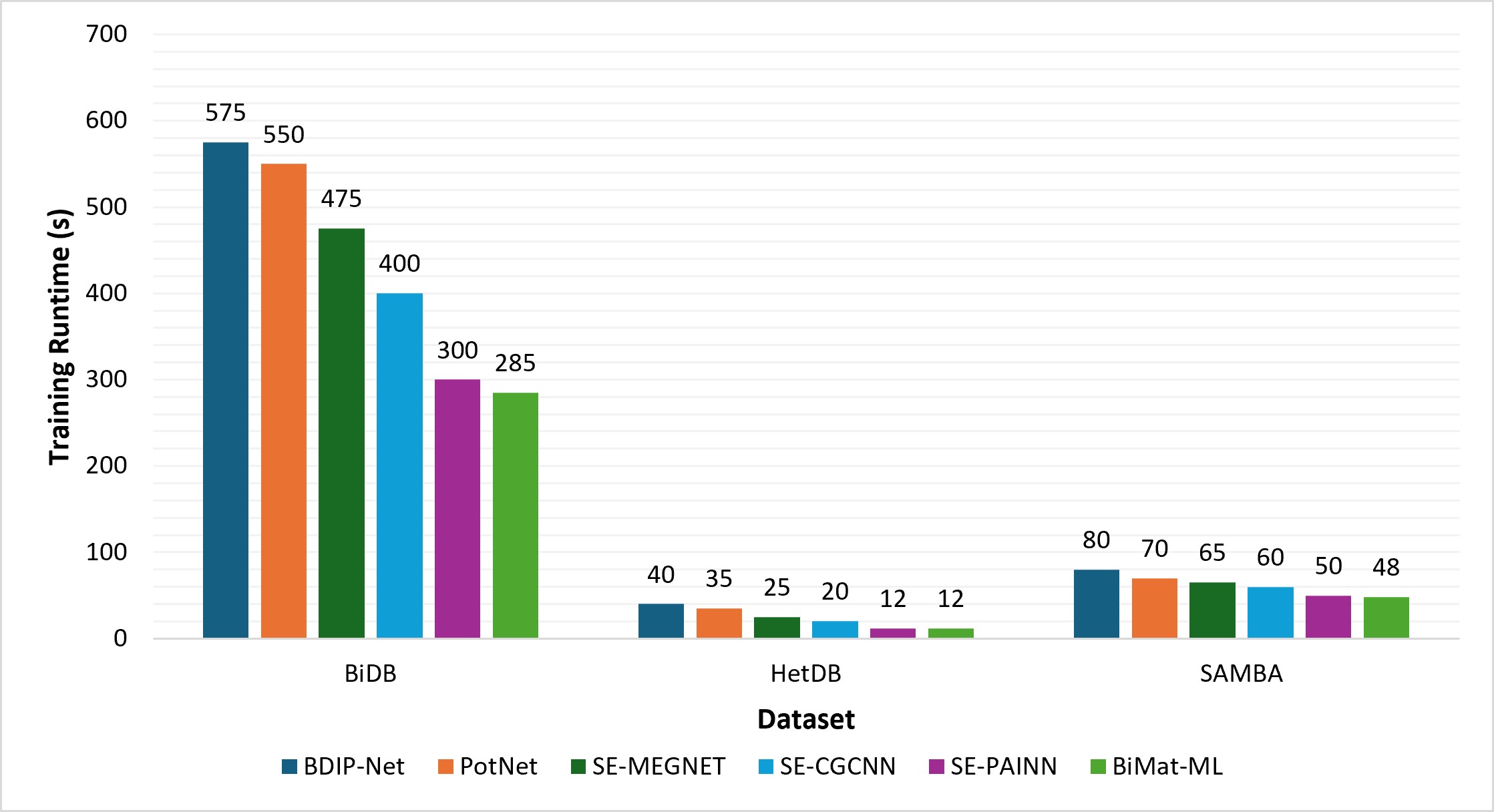}
    \caption{ Training Runtime Comparison Across 3 Datasets}
    \label{fig:training_time}
\end{figure}
We compare the training and inference runtimes of \mbox{BDIP-Net}, PotNet and the baseline models on BiDB, HetDB, and SAMBA. Since the training and inference runtimes are nearly identical for structures optimized by \mbox{DFT-PBE-D3} and \mbox{MatterSim-D3}, we report a single runtime for each dataset and model. The training runtimes are summarized in Figure~\ref{fig:training_time}. On BiDB, \mbox{BDIP-Net} requires \(575~\mathrm{s}\) for training, which is comparable to PotNet (\(550~\mathrm{s}\)) and SE-MEGNET (\(475~\mathrm{s}\)), but slower than SE-CGCNN (\(400~\mathrm{s}\)), SE-PAINN (\(300~\mathrm{s}\)), and BiMat-ML (\(285~\mathrm{s}\)). On HetDB, \mbox{BDIP-Net} requires \(40~\mathrm{s}\), compared with \(35~\mathrm{s}\) for PotNet, \(25~\mathrm{s}\) for SE-MEGNET, \(20~\mathrm{s}\) for SE-CGCNN, and \(12~\mathrm{s}\) for both SE-PAINN and BiMat-ML. On SAMBA, \mbox{BDIP-Net} requires \(80~\mathrm{s}\), compared with \(70~\mathrm{s}\) for PotNet, \(65~\mathrm{s}\) for SE-MEGNET, \(60~\mathrm{s}\) for SE-CGCNN, \(50~\mathrm{s}\) for SE-PAINN, and \(48~\mathrm{s}\) for BiMat-ML. Across all datasets and models, the average per-sample inference time is approximately \(0.05~\mathrm{s}\). Although the training and inference runtimes are nearly identical when using \mbox{MatterSim-D3}-optimized and \mbox{DFT-PBE-D3}-optimized structures as model inputs, the time required to obtain these two types of input structures is significantly different. Starting from the same initial bilayer structure, \mbox{MatterSim-D3} can generate a structurally optimized bilayer structure in only a few seconds, whereas obtaining a \mbox{DFT-PBE-D3}-optimized structure typically requires hours of DFT-based structural optimization. Thus, \mbox{BDIP-Net} combined with \mbox{MatterSim-D3}-based structure generation provides a computationally practical pipeline for bilayer property prediction.

\subsection{Cross-Domain Generalization}

\subsubsection{Monolayer Overlap}
\label{app:cross_domain_mono_overlap}

Table~\ref{tab:monolayer_list} lists the 63 unique monolayers that appear in the SAMBA dataset. Monolayers overlapping with the BiDB dataset are marked with the subscript \((B)\), while those overlapping with both the BiDB and HetDB datasets are marked with \((B,H)\). This annotation is used to identify monolayer overlap between the training and test datasets in the cross-domain generalization analysis presented in the main paper.

\begin{table}[t]
\centering
\caption{List of monolayers in the SAMBA dataset. Subscripts
\((B)\) and \((B,H)\) indicate monolayers overlapping with BiDB
only and with both BiDB and HetDB, respectively.}
\label{tab:monolayer_list}
\renewcommand{\arraystretch}{1.15}
\setlength{\tabcolsep}{6pt}

\newcommand{\bmark}[1]{#1_{\scriptscriptstyle(B)}}
\newcommand{\bhmark}[1]{#1_{\scriptscriptstyle(B,H)}}

\resizebox{\columnwidth}{!}{%
\begin{tabular}{ccccc}
\toprule

$\mathrm{As_2}$ &
$\mathrm{As_2S_3}$ &
$\mathrm{As}$ &
$\bmark{\mathrm{Bi_2Te_3}}$ &
$\mathrm{As_4S_6^{Orpiment}}$ \\

$\mathrm{Bi_2}$ &
$\mathrm{Bi_2Se_3}$ &
$\bmark{\mathrm{Bi_2SeTe_2}}$ &
$\bhmark{\mathrm{BN}}$ &
$\mathrm{As_4S_6^{Anorpiment}}$ \\

$\mathrm{BP}$ &
$\bhmark{\mathrm{C_2}}$ &
$\mathrm{CdTe}$ &
$\bhmark{\mathrm{Ga_2S_2}}$ &
$\bhmark{\mathrm{Ga_2Se_2}}$ \\

$\bmark{\mathrm{Ga_2Te_2}}$ &
$\mathrm{Ge_2}$ &
$\bmark{\mathrm{GeS}}$ &
$\mathrm{Hf_2Te_6}$ &
$\bmark{\mathrm{HgTe}}$ \\

$\bhmark{\mathrm{In_2Se_2}}$ &
$\mathrm{In_2Se_3}$ &
$\mathrm{IrTe_2^{1T}}$ &
$\bhmark{\mathrm{MoS_2^{2H}}}$ &
$\bhmark{\mathrm{MoSe_2^{2H}}}$ \\

$\bhmark{\mathrm{MoTe_2^{2H}}}$ &
$\bmark{\mathrm{NbS_2^{2H}}}$ &
$\bmark{\mathrm{NbSe_2^{2H}}}$ &
$\bmark{\mathrm{NbTe_2^{2H}}}$ &
$\bmark{\mathrm{NiS_2^{1T}}}$ \\

$\bmark{\mathrm{NiSe_2^{1T}}}$ &
$\bmark{\mathrm{NiTe_2^{1T}}}$ &
$\mathrm{P_2}$ &
$\mathrm{P_4}$ &
$\bmark{\mathrm{Pd_2Se_4}}$ \\

$\mathrm{ReSe_2^{2H}}$ &
$\bmark{\mathrm{PdS_2^{1T}}}$ &
$\bmark{\mathrm{PdSe_2^{1T}}}$ &
$\bmark{\mathrm{PdTe_2^{1T}}}$ &
$\mathrm{Pd_4Se_6Hg_2}$ \\

$\bmark{\mathrm{PtS_2^{1T}}}$ &
$\bhmark{\mathrm{PtSe_2^{1T}}}$ &
$\bmark{\mathrm{PtTe_2^{1T}}}$ &
$\mathrm{Pt_4Se_6Hg_2}$ &
$\mathrm{ReS_2^{2H}}$ \\

$\mathrm{Sb_2}$ &
$\bmark{\mathrm{Sb_2Te_3}}$ &
$\mathrm{Si_2}$ &
$\bmark{\mathrm{SnS_2^{1T}}}$ &
$\bmark{\mathrm{SnS}}$ \\

$\bmark{\mathrm{SnSe_2^{1T}}}$ &
$\bmark{\mathrm{TaS_2^{2H}}}$ &
$\bmark{\mathrm{TaSe_2^{2H}}}$ &
$\mathrm{Ti_2S_6}$ &
$\mathrm{TiS_2^{1T}}$ \\

$\bmark{\mathrm{TiSe_2^{1T}}}$ &
$\bhmark{\mathrm{WS_2^{2H}}}$ &
$\bhmark{\mathrm{WSe_2^{2H}}}$ &
$\bhmark{\mathrm{WTe_2^{2H}}}$ &
$\bmark{\mathrm{Zn_2Se_2}}$ \\

$\mathrm{Zr_2Te_6}$ &
$\bmark{\mathrm{ZrS_2^{1T}}}$ &
$\bhmark{\mathrm{ZrSe_2^{1T}}}$ &
&
\\

\bottomrule
\end{tabular}%
}
\end{table}

\subsubsection{HetDB-to-SAMBA}
\label{app:hetdb_to_samba}
In this setting, the models are trained on the full HetDB dataset consisting of 336 heterobilayers and evaluated on the full SAMBA test set of 980 bilayers. Among the monolayers forming the HetDB heterobilayers, 13 overlap with monolayers in SAMBA. Both BDIP-Net and PotNet are trained with a learning rate of \(2\times10^{-4}\) and a batch size of 128, while their remaining model configurations follow those described in Section~\ref{sec:exp_setting}. Similar to the previous setting, the SAMBA test set is divided according to the degree of monolayer overlap with the training data. The no-monolayer-overlap, one-monolayer-overlap, and two-monolayer-overlap subsets contain 626, 284, and 70 bilayers, respectively. The two-monolayer-overlap subset is further divided into 36 exact-pair-overlap bilayers and 34 bilayers with two-monolayer overlap excluding exact-pair overlap. Table~\ref{tab:cross_domain_generalization2} summarizes the prediction performance of BDIP-Net and PotNet under this setting.

\begin{table}[t]
\centering
\caption{HetDB-to-SAMBA domain generalization prediction performance of BDIP-Net and PotNet.}
\label{tab:cross_domain_generalization2}
\renewcommand{\arraystretch}{1.15}

\resizebox{\columnwidth}{!}{%
\begin{tabular}{lclcccc}
\toprule
\textbf{Setting} &
\textbf{\makecell{Number of\\Bilayers}} &
\textbf{Model} &
\textbf{MAE} $\downarrow$ &
\textbf{MSE} $\downarrow$ &
\textbf{RMSE} $\downarrow$ &
\textbf{R$^2$} $\uparrow$ \\
\midrule

\multirow{2}{*}{Full SAMBA}
& \multirow{2}{*}{980}
& BDIP-Net & 0.56 & 0.50 & 0.71 & -0.93 \\

&
& \textbf{PotNet} & \textbf{0.50} & \textbf{0.39} & \textbf{0.63} & \textbf{-0.50} \\
\midrule

\multirow{2}{*}{No-monolayer-overlap}
& \multirow{2}{*}{626}
& BDIP-Net & 0.65 & 0.61 & 0.78 & -2.71 \\

&
& \textbf{PotNet} & \textbf{0.57} & \textbf{0.46} & \textbf{0.68} & \textbf{-1.77} \\
\midrule

\multirow{2}{*}{One-monolayer-overlap}
& \multirow{2}{*}{284}
& BDIP-Net & 0.46 & 0.38 & 0.61 & -0.85 \\

&
& \textbf{PotNet} & \textbf{0.43} & \textbf{0.32} & \textbf{0.56} & \textbf{-0.56} \\
\midrule

\multirow{2}{*}{Two-monolayer-overlap}
& \multirow{2}{*}{70}
& \textbf{BDIP-Net} & \textbf{0.12} & \textbf{0.03} & \textbf{0.16} & \textbf{0.97} \\

&
& PotNet & 0.15 & 0.10 & 0.31 & 0.89 \\
\midrule

\multirow{2}{*}{Exact-pair-overlap}
& \multirow{2}{*}{36}
& BDIP-Net & 0.10 & 0.02 & 0.14 & 0.95 \\

&
& \textbf{PotNet} & \textbf{0.07} & \textbf{0.01} & \textbf{0.11} & \textbf{0.97} \\
\midrule

\multirow{2}{*}{\makecell[l]{Two-monolayer-overlap excluding\\exact-pair-overlap}}
& \multirow{2}{*}{34}
& \textbf{BDIP-Net} & \textbf{0.14} & \textbf{0.03} & \textbf{0.18} & \textbf{0.98} \\

&
& PotNet & 0.24 & 0.19 & 0.43 & 0.86 \\
\bottomrule

\end{tabular}%
}
\end{table}

\begin{figure}[H]
    \centering
    \includegraphics[width=1\linewidth]{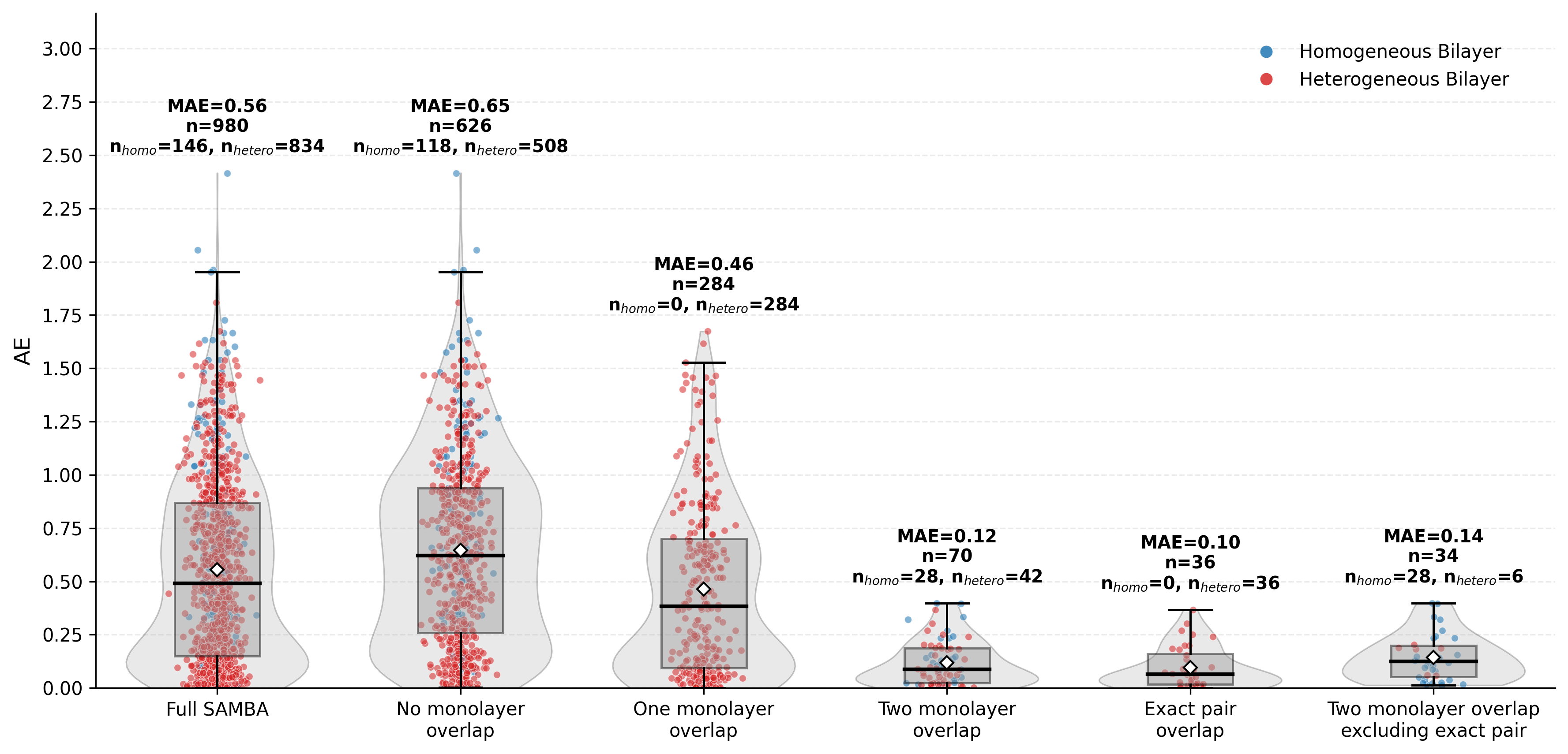}
    \caption{HetDB-to-SAMBA Domain Generalization Breakdown Analysis}
    \label{fig:box_plot_hetdb_samba_mono}
\end{figure}

\begin{figure}[H]
    \centering
    \includegraphics[width=1\linewidth]{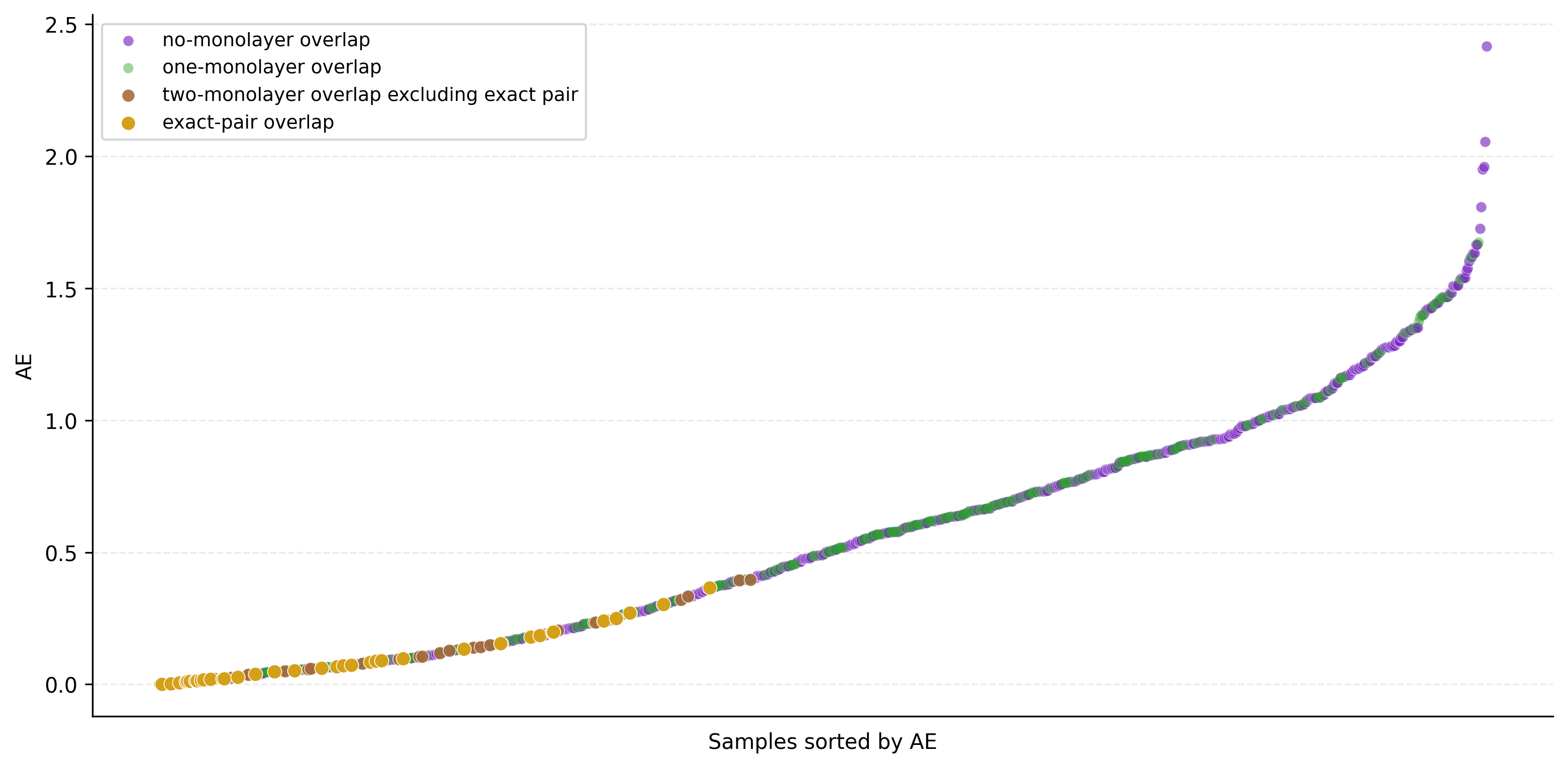}
    \caption{HetDB-to-SAMBA Domain Generalization AE Distribution}
    \label{fig:scatter_plot_hetdb_samba_mono}
\end{figure}

\noindent\textbf{Prediction Performance.}
As shown in Table~\ref{tab:cross_domain_generalization2}, PotNet achieves better prediction performance than BDIP-Net on the full SAMBA test set of 980 bilayers, with an MAE of \(0.50\), an MSE of \(0.39\), an RMSE of \(0.63\), and an \(R^2\) of \(-0.50\), compared with \(0.56\), \(0.50\), \(0.71\), and \(-0.93\), respectively, for BDIP-Net. Figs.~\ref{fig:box_plot_hetdb_samba_mono} and~\ref{fig:scatter_plot_hetdb_samba_mono} show that the prediction performance of both models improves as the degree of monolayer overlap increases. On the 626 no-monolayer-overlap bilayers, PotNet achieves a lower MAE than BDIP-Net (\(0.57\) vs.\ \(0.65\)). On the 284 one-monolayer-overlap bilayers, the corresponding MAEs are \(0.43\) and \(0.46\). In contrast, BDIP-Net performs better on the 70 two-monolayer-overlap bilayers, achieving an MAE of \(0.12\), compared with \(0.15\) for PotNet. The best results are obtained on the 36 exact-pair-overlap bilayers, where PotNet achieves a slightly lower MAE (\(0.07\) vs.\ \(0.10\)). However, after removing these exact-pair-overlap bilayers, BDIP-Net achieves an MAE of \(0.14\), an RMSE of \(0.18\), and an \(R^2\) of \(0.98\) on the remaining 34 two-monolayer-overlap bilayers, compared with \(0.24\), \(0.43\), and \(0.86\), respectively, for PotNet. These results indicate that although PotNet performs better overall, BDIP-Net provides stronger cross-dataset generalization to unseen bilayer combinations of known monolayers. The overall HetDB-to-SAMBA performance is nevertheless limited because most SAMBA bilayers belong to the no-monolayer-overlap and one-monolayer-overlap subsets.

\end{document}